\documentclass{article}

\usepackage[final]{neurips_2024}

\usepackage{natbib}
\setcitestyle{numbers,square}
\usepackage[utf8]{inputenc} % allow utf-8 input
\usepackage[T1]{fontenc}    % use 8-bit T1 fonts
\usepackage{hyperref}       % hyperlinks
\usepackage{url}            % simple URL typesetting
\usepackage{booktabs}       % professional-quality tables
\usepackage{amsfonts}       % blackboard math symbols
\usepackage{nicefrac}       % compact symbols for 1/2, etc.
\usepackage{microtype}      % microtypography
\usepackage{xcolor}         % colors
\usepackage{colortbl}
\usepackage{color}
\hypersetup{
    colorlinks=true,
    linkcolor=blue,
    filecolor=blue,      
    urlcolor=blue,
    citecolor=cyan,
}
\usepackage{cite}
\usepackage{times}
\usepackage{latexsym}
\usepackage{array}
\usepackage{graphicx}
\usepackage{epstopdf}
\usepackage{makecell}
\newcolumntype{L}[1]{>{\raggedright\let\newline\\\arraybackslash\hspace{0pt}}m{#1}}
\newcolumntype{C}[1]{>{\centering\let\newline\\\arraybackslash\hspace{0pt}}m{#1}}
\newcolumntype{R}[1]{>{\raggedleft\let\newline\\\arraybackslash\hspace{0pt}}m{#1}}
\usepackage{multirow}
\usepackage{bbding}
\usepackage{pifont}
\usepackage{amsmath}
\usepackage{arydshln}
\usepackage{cleveref}
\usepackage{wrapfig}
\usepackage{enumitem}
\crefname{section}{§}{§§}
\Crefname{section}{§}{§§}

\title{From Instance Training to Instruction Learning:\\ Task Adapters Generation from Instructions}
\author{Huanxuan Liao$^{1,2}$, Shizhu He$^{1,2}$\thanks{Corresponding author} , Yao Xu$^{1,2}$, Yuanzhe Zhang$^{1,2}$, \\ \textbf{Yanchao Ha}o$^{3}$, \textbf{Shengping Liu}$^{4}$, \textbf{Kang Liu}$^{1,2}$, \textbf{Jun Zhao}$^{1,2}$  \\
    $^1$ The Key Laboratory of Cognition and Decision Intelligence for Complex Systems, \\
    Institute of Automation, Chinese Academy of Sciences, Beijing, China \\
    $^2$ School of Artificial Intelligence, University of Chinese Academy of Sciences, Beijing, China \\
    $^3$ Platform and Content Group, Tencent, Beijing, China ~~~
    $^4$ Unisound, Beijing, China\\
  {liaohuanxuan2023@ia.ac.cn} {\{shizhu.he, yao.xu, kliu, jzhao\}@nlpr.ia.ac.cn} \\}
\begin{document}

\maketitle

\begin{abstract}
  Large language models (LLMs) have acquired the ability to solve general tasks by utilizing instruction finetuning (IFT).
  %, which describes different tasks in the same format of natural language
    However, IFT still relies heavily on instance training of extensive task data, which greatly limits the adaptability of LLMs to real-world scenarios where labeled task instances are scarce and broader task generalization becomes paramount. 
    Contrary to LLMs, humans acquire skills and complete tasks not merely through repeated practice but also by understanding and following instructional guidelines. This paper is dedicated to simulating human learning to address the shortcomings of instance training, focusing on instruction learning to enhance cross-task generalization. Within this context, we introduce \textbf{T}ask \textbf{A}dapters \textbf{G}eneration from \textbf{I}nstructions (\textbf{TAGI}), which automatically constructs the task-specific model in a parameter generation manner based on the given task instructions without retraining for unseen tasks. 
    Specifically, we utilize knowledge distillation to enhance the consistency between TAGI developed through \textit{Learning with Instruction} and task-specific models developed through \textit{Training with Instance}, by aligning the labels, output logits, and adapter parameters between them. TAGI is endowed with cross-task generalization capabilities through a two-stage training process that includes hypernetwork pretraining and finetuning.
   We evaluate TAGI on the Super-Natural Instructions and P3 datasets. %highlighting its exceptional proficiency in generating parameters for unforeseen tasks. 
   The experimental results demonstrate that TAGI can match or even outperform traditional meta-trained models and other hypernetwork models, while significantly reducing computational requirements. Our code will be available at \url{https://github.com/Xnhyacinth/TAGI}.
\end{abstract}

\section{Introduction}
\label{introduction}
Large language models (LLMs) have acquired the ability to solve general tasks by utilizing instruction finetuning (IFT), which describes different tasks in the same natural language format \citep{gpt3, scaling, naturalinstructions}. However, IFT still relies heavily on instance training of extensive task data \{(\textit{Description}, \textit{[Demostrations]}, \textit{Source}, \textit{Target})\} \citep{super, learning}, which faces significant limitations in adapting LLMs to real-world scenarios where labeled task instances are scarce and broader task generalization becomes paramount.

\begin{wrapfigure}[16]{r}{0.5\textwidth}
\vspace{-0.4cm}
\centering
\includegraphics[width=0.5\textwidth]{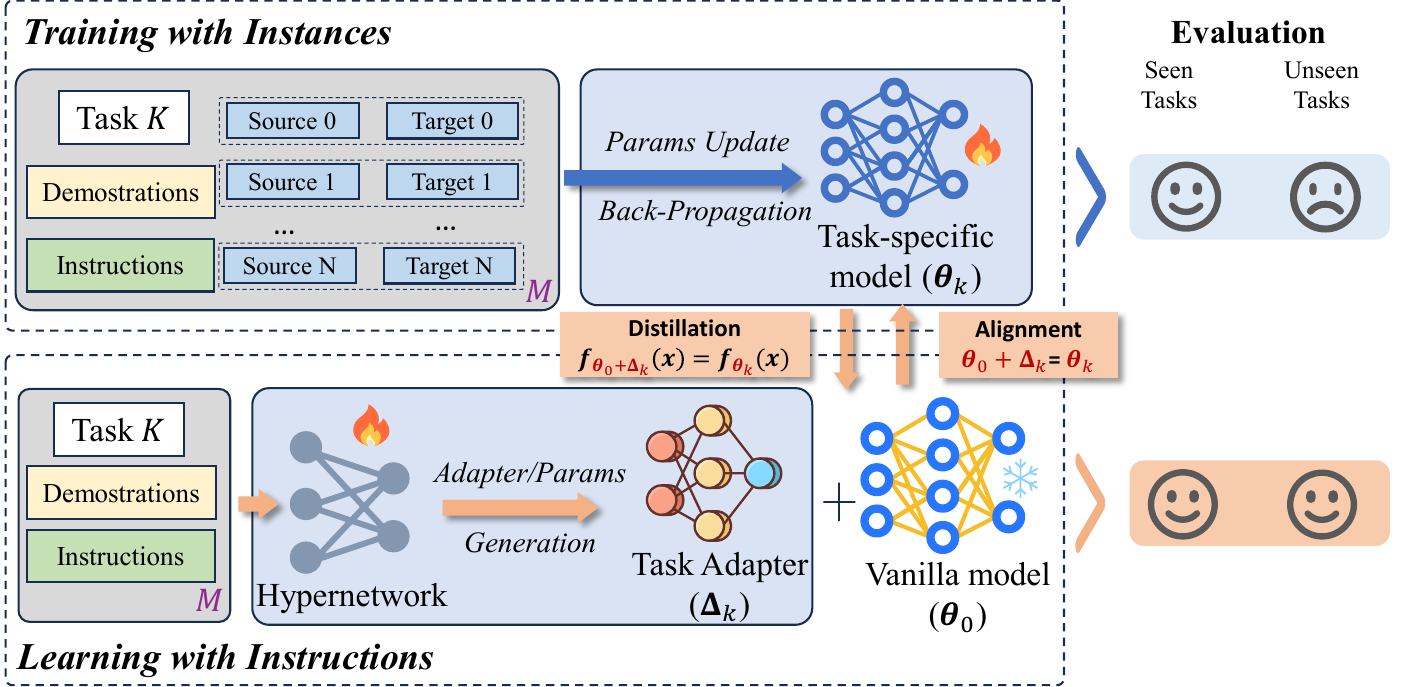}
% \vspace{-0.45cm}
\caption{Comparison of the typical \textbf{Training with Instance} and the proposed \textbf{Learning with Instruction}: The former involves training the model at the instance level with parameter updates, while the latter generates a task-specific adapter at the task level with parameter generation.
}
\label{intro}
% \vspace{-0.7cm}
\end{wrapfigure}

Therefore, for better cross-task generalization, the "zero-shot" learning ability of LLMs is crucial for real-world applications: models learned with instructions can achieve non-trivial performance on unseen tasks with just a single instruction that provides a comprehensive description of the task
% Therefore, the "zero-shot" learning ability of LLMs is crucial for their application in real-world scenarios: models learned with instructions can achieve non-trivial performance on a given unseen task with just a single instruction that provides a comprehensive description of the task %or outlines the precise steps required for its successful execution
(e.g., "\textit{You will be given sentences in which your task is to recognize the name of a person.}"). Traditionally, achieving this capability involves meta-training the model by associating each input with specific task instructions \citep{metaicl, super}.
For example, GPT-3 \citep{training} has demonstrated strong "zero-shot" capabilities through meta-training. However, these methods heavily depend on the foundation model's abilities and are inefficient for various unseen tasks \citep{reframing, adapting}, as they require reprocessing extensive task instructions and some supplementary task data (e.g., examples from few-shot instances) for each input (see the top of \autoref{intro}).
 
%Moreover, they are constrained either to tasks mirroring the format of their training counterparts (e.g., question answering format) \citep{adapting} or to those heavily dependent on task-specific templates \citep{flan}, where minor modifications can precipitate substantial performance disparities \citep{reframing}. 

In recent years, researchers have begun to explore meta-learning to enhance the cross-task generalization capabilities of LLMs, aiming to construct flexible, reusable and robust task-specific models \citep{meta-learn, LearningTL}. For example, task-specific models such as Adapter \citep{houlsby2019parameterefficient}, LoRA \citep{hu2022lora}, and Prefix \citep{prefix} have been constructed by a hypernetwork \citep{hypernetworks}. This approach significantly enhances task generalization by processing instructions efficiently, reducing redundant computations \citep{hypertuning}. However, these methods heavily depend on a substantial corpus of training instances, which can hinder their capacity to efficiently learn and construct task-specific models based on provided instructions \citep{hint}.

In fact, contrary to LLMs, humans acquire skills and complete tasks not only through repeated practice but also by understanding and following instructional guidelines \citep{teach}. For example, a tourist with basic knowledge of riding vehicles can easily learn to use new ones abroad for the first time with the help of travel guides.
%For example, a tourist who has mastered the basic knowledge of riding vehicles can easily learn how to take new vehicles even if it is her/his first time to travel abroad, merely with the help of travel guides.
This paper aims to mimic the way humans learn skills by understanding instructions. This shift represents a modest evolution in task model construction, transitioning from traditional instance training models to a contemporary approach focused on instruction learning.
%This shift represents a modest evolution in the current task model construction paradigm, moving away from traditional instance training models towards a more contemporary approach centered on instruction learning. 
By providing task instructions, the novel paradigm offers an automated solution for generating task-specific adapters and seamlessly integrating them into the base model. This approach aims to streamline the development of task-specific models while enhancing their ability to generalize across diverse tasks with instructions.

Guided by this goal, we introduce \textbf{T}ask \textbf{A}dapters \textbf{G}eneration from \textbf{I}nstructions (TAGI), which converts instructions to task-specific adapters using a hypernetwork. Under the knowledge distillation framework \citep{kd, minilm}, we enable models to the "\textit{Learning with Instruction}" paradigm in a manner analogous to the "\textit{Training with Instance}" paradigm. TAGI will enhance the alignment between the task-specific model $\theta_k$ (acting as the teacher) and the vanilla LLM $\theta_0$ combined with the generated task adapters $\Delta_k$ (acting as the student) (see the bottom of \autoref{intro}). This alignment is achieved not only through instance training but also by incorporating parameter learning for task-specific models based on instructions. Specifically, we align the student under two distinct paradigms, encompassing not just the targets and logits, but also the adapters' parameters by an L2 regularization within instruction, which represents the enhancement of the understanding of instructions and the ability to generate more efficient task-specific adapters. Moreover, TAGI endows the model with task generalization capabilities through a two-stage training process: hypernetwork pretraining on standard text pretraining data (e.g., C4 \citep{t5}), followed by finetuning on meta-training tasks. This allows it to generalize effectively across unseen tasks without sacrificing performance.

We evaluate TAGI on the Super-Natural Instructions (SNI) \citep{super} and P3 \citep{multitask} datasets. Experimental results demonstrate its ability to effectively generate adapters for unseen tasks, surpassing meta-trained models by 2\% in SNI and 5\% in P3, while significantly reducing computational demands by 60\%, and outperforming other hypernetwork models by 7\%.
%We find that TAGI outperforms strong baselines in both zero-shot and few-shot settings, which 
% It validates our hypothesis that learning from instructions can be achieved by leveraging supervision from the parameters of the target task-specific model from LoRA tuning. Notably, our method does not necessitate the addition of extra model parameters or gradients, and it circumvents the inefficiency of encoding task instructions repeatedly during testing, thereby offering a significant trade-off between performance and efficiency. We summarize our contributions as follows,
% It validates our hypothesis that learning from instructions can be achieved by leveraging supervision from the parameters of the target task-specific model obtained through LoRA tuning. 
Notably, our method does not require additional parameter updating or gradient back-propagation, and it avoids the inefficiency of repeatedly encoding instructions during inference. 
% offering a significant trade-off between performance and efficiency. 
We summarize our contributions as follows:
% \noindent \textbf{Contribution.} We summarize our contributions as follows,
% \vspace{-0.4cm}
\begin{itemize}[leftmargin=*]
    % \vspace{-0.4cm}
    \item We propose a novel model construction paradigm by imitating human learning abilities, \textbf{Learning with Instruction}, for the cross-task generalization of the LLMs.  To the best of our knowledge, it is the first time that a task-specific model has been generated based on instruction learning, and its capabilities and parameters are distilled from a teacher model trained on instance learning.
    % \vspace{-0.1cm}
    \item We used a knowledge distillation framework to develop task-specific models within the instruction learning paradigm. By aligning model parameters comprehensively, the TAGI method improves the model's ability to understand instructions and solve unseen tasks more accurately and efficiently.
    % \vspace{-0.2cm}
    \item Comprehensive quantitative and qualitative assessments have highlighted the effectiveness of TAGI on two publicly available large-scale instruction datasets, with lower inference costs.
\end{itemize}

% \vspace{-1cm}
\section{Related Work}

TAGI draws inspiration from previous research on instruction following, hypernetworks and knowledge distillation. In this section, we will delve into the pioneering work in these areas.

\noindent \textbf{Instruction Following} is often used to evaluate the cross-task generalization of LLMs, and it is dedicated to handling any task described in natural language. Recent findings suggest that additional finetuning of LLMs with instructions substantially improves their zero-shot capabilities \citep{scaling, flan, learning}. 
% This enhancement allows the models to adeptly utilize instructions for task execution. 
Moreover, large-scale multi-task meta-training has been shown to equip models with the ability to address new tasks in zero- or few-shot scenarios, facilitated by standard task formats and prompts \citep{multitask, adapting} alongside providing concise task instructions and select examples \citep{crosstask, super}. However, the instructions and examples can significantly escalate the computational burden compared to task-specific models. Existing works attempt to mitigate this issue involved creating adapters to separately process instructions and examples \citep{hint, ye2021learning} with reduced performance. To overcome these limitations, we introduce a new paradigm that draws on instruction-based learning, simulating instance training to enhance the perception and processing capabilities of LLMs for handling unseen tasks.

\noindent \textbf{Hypernetworks} \citep{hypernetworks, Schmidhuber1992LearningTC} are neural networks that generate weights for other neural networks \citep{brief}, which are designed to use fewer parameters to dynamically build task-specific models \citep{hyperprompt, hypergrid}. Notable works such as HyperTuning \citep{hypertuning}, HINT \citep{hint}, and Hypter \citep{ye2021learning} have all adopted hypernetworks to convert task instructions and demonstrations into adapters for LLMs. And MEND \citep{demonstration} utilizes hypernetworks to compress demonstrations for distilled vectors. Although they all avoided processing lengthy instructions repeatedly and utilized adapters to make training and testing more cost-effective \citep{fewshot}, they still have a performance loss compared to meta-training \citep{deb2022boosting}. The proposed method TAGI incorporates the utilization of hypernetworks, which are instrumental in generating task-specific adapters that are seamlessly integrated into LLMs. Compared to existing models based on hypernetworks, TAGI not only trains at the instance level but also incorporates knowledge distillation to supervise the adapters generated by hypernetworks, thereby achieving both efficiency and effectiveness.

\noindent \textbf{Knowledge Distillation} is a technique where a smaller model (student) learns to mimic the predictions of a larger model (teacher), aiming to retain performance while reducing computational resources \citep{kd}. Indeed, the application of knowledge distillation is the essential difference between the proposed method in this paper and other hypernetwork-based methods such as HINT \citep{hint} and Hypter \citep{ye2021learning}. Recently, some works \citep{learningbydistill} utilize knowledge distillation to finetune small language models such as T5 \citep{t5}, enabling them to act as LLMs with pre-prompting without any given prompts. Compared with the typical knowledge distillation methods of LLMs, the proposed method TAGI in this paper further utilizes model parameter alignment and aims to mimic another learning paradigm of human skill learning. We not only calculate the Kullback–Leibler (KL) divergence \citep{KL} between teacher and student models \citep{kd}, but also compute the L2 regularization between the generated adapter by instruction learning and task-specific models by instance training.

\section{Methods}
\subsection{Problem Setting}
\noindent \textbf{Cross-task Generalization:} Given a set of tasks $\mathcal{T} = \{ \mathcal{T}_1, ... , \mathcal{T}_{|\mathcal{T}|} \} $, where each task $\mathcal{T}_i$ contains a set of (\textit{source}, \textit{target}) samples $\mathcal{D}_i = \{ (s_1, t_1), ... , (s_n, t_n) \} $. We categorize these tasks into three distinct non-overlapping groups for validating out-of-distribution generalization: meta-train ($\mathcal{T}_{train}$), meta-valid ($\mathcal{T}_{valid}$), and meta-test ($\mathcal{T}_{test}$), assuming all tasks adhere to a text-to-text format. For example, $\mathcal{T}_{train}$ comprises tasks like translation and question answering, the $\mathcal{T}_{valid}$ and $\mathcal{T}_{test}$ encompass tasks such as paraphrasing and natural language inference respectively.
 Within the $\mathcal{T}_{train}$, the goal is to utilize the data for training and transfer knowledge to facilitate learning to resolve the test tasks. For all methods discussed, aside from the original unsupervised pretraining of the language model backbone on separate corpora, the model learning primarily takes place through multi-task training on the $\mathcal{T}_{train}$.

\subsection{Task Adapters Generation from Instructions (TAGI)}
In this section, we will introduce the detailed method of TAGI. For each (unseen) task, TAGI consists of two core components: a \textbf{hypernetwork} \cref{hyper} which receives task instructions and generates parameter-efficient adapters, and a task-specific model which combines the \textbf{vanilla LLM} and the generated adapters from hypernetwork. 

\begin{figure}[t]
\centerline{\includegraphics[width=1.0\textwidth]{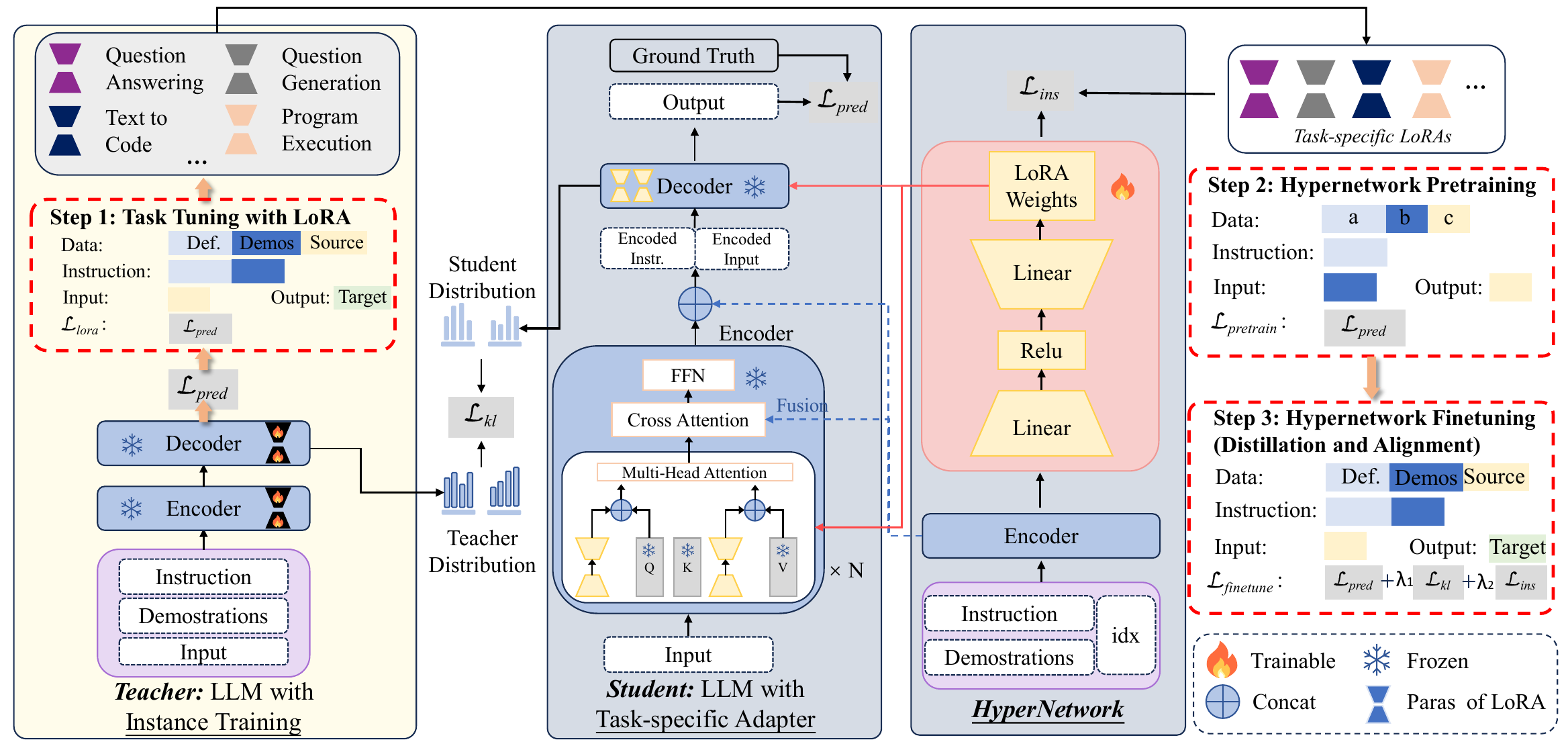}}
% \vspace{-0.45cm}
\caption{Overview of TAGI. The hypernetwork takes instruction as input and generates adapters subsequently integrated into the vanilla LLM, and constructed the task-specific model as student. After training the task models through instances on multiple basic tasks as a teacher, TAGI constructs task-specific models by aligning the labels, output logits, and adapter parameters between teacher and student models. To improve compliance with task instructions and the efficacy of weight generation, TAGI undergoes a two-stage hypernetwork training process: hypernetwork pretraining and finetuning. a-c are random divisions of the sampled sentences from pretraining datasets.
}
\label{model}
% \vspace{-0.6cm}
\end{figure}

Unlike traditional meta-training methods, we transition from \textit{training with instance} to \textit{learning with instruction}, which not only addresses efficiency issues at the instance level but also incorporates parameter alignment for the task-specific model parameters at the instruction level. 
Specifically, the complete process is shown in \autoref{model}, we initially train the LoRA modules \cref{lora} on various upstream tasks (seen tasks) with task datasets of meta-train ($\mathcal{T}_{train}$). Specifically, for $N$ distinct upstream tasks, we independently train $N$ LoRA modules, with each module denoted as $\Delta_i$ for task $\mathcal{T}_i \in \mathcal{T}$, presumed to represent the optimal model for its respective task. Subsequently, TAGI is committed to building proprietary models for downstream tasks (unseen tasks). Its training process is bifurcated into two primary phases: hypernetwork pretraining \cref{pretraining} and hypernetwork finetuning \cref{finetuning} which encompasses distillation and alignment. 

% This makes the model more sensitive to task instructions and more accurate in generating model parameters to solve tasks, while also leveraging knowledge distillation to incorporate implicit learning of strong baselines.

\subsubsection{Hypernetwork for Converting Instructions into LoRA}
\label{hyper}

A pivotal element of our model is the hypernetwork that converts task instructions (descriptions and demonstrations) into a parameter-efficient module. Our hypernetwork comprises two crucial components: the \textbf{encoder}, derived from the vanilla LLM\footnote{We find that re-using the encoder from the vanilla LLM works well \citep{hint}.}, is designed to minimize encoding biases by converting task instructions into a continuous contextual representation. This representation is then fused with LLM input and concated with encoded input for the decoder. Additionally, the \textbf{adapter generator}, utilizing a basic MLP design \citep{liao2025awakening}, is both lightweight and efficient, effectively converting encoded instructions into parameter-efficient modules.

\noindent \textbf{Encoder:} Prior studies simply concatenated encoded instructions with inputs, overlooking the interactions between them. To address this, we integrated a hierarchical cross-attention layer into the encoder of the LLM to refine the input representation with embedded instruction details. Specifically, for an input \(x\) and its corresponding task instruction \(i_{x}\), we initially employ the encoder within the hypernetwork to encode the instruction into representations $\textbf{I}_{x} \in \mathbb{R}^{s \times d}$. Then, we feed the \(x\) into the model and obtain the output representation $\textbf{S}_{l}$ from the self-attention sublayer in the \(l\)-th layer. Ultimately, $\textbf{S}_{l}$ is processed through the \(l\)-th cross-attention layer, resulting in a text representation that is enriched with instruction information:
\begin{equation}
   \textbf{F}_{l} = \text{CrossAttentionLayer}_l(\textbf{S}_{l}, \textbf{I}_{x})
\end{equation}
where $\text{CrossAttentionLayer}_l$ conducts multi-head attention on the query, key, and value matrices, followed by residual connection and layer normalization. The final input to the decoder is the concatenation of the encoded instruction and the encoded fusion input, i.e., $(\textbf{I}_{x}; \textbf{F}_{l})$. %where $;$ is the concatenation.

\noindent \textbf{Adapter Generator:} Considering the efficiency and effectiveness, we utilize a two-layer multi-layer perceptron (MLP) to generate parameter-efficient modules (e.g., LoRA) for the encoded instruction. To differentiate between the query $\mathcal{Q}$ and value $\mathcal{V}$ matrices as well as the layers, we introduce layer ids $\text{idx}_l^{\{\mathcal{Q}, \mathcal{V}\}} \in \{0, \ldots, 2 \times \# \text{blocks}\}$ as positional information. We use a unique network for each layer and share it between $\mathcal{Q}$ and $\mathcal{V}$ (i.e., one network is used for a certain layer LoRA generation).
\begin{equation}
   \text{LoRA}_{l}^{\{\mathcal{Q}, \mathcal{V}\}} = \text{MLP}_l(\textbf{I}_{x_k}; \text{idx}_l^{\{\mathcal{Q}, \mathcal{V}\}} ~|~ \text{idx}_l^\mathcal{Q} = 2l, \text{idx}_l^\mathcal{V} = 2l+1)
\end{equation}
where $\text{LoRA}_{l}^{\mathcal{Q}}$ and $\text{LoRA}_{l}^{\mathcal{V}}$ are the $l$-th LoRA of $\mathcal{Q}$ and $\mathcal{V}$, respectively.

\subsubsection{LoRA Tuning for Task-specific Models}
\label{lora}
LoRA \citep{hu2022lora} efficiently reduces the number of trainable parameters by decomposing the update of the LLM’s attention weight matrix (denoted as \(\mathbf{W}_0 \in \mathbb{R}^{d \times k}\)) into low-rank matrices. Specifically, LoRA updates the weight matrix as \(\mathbf{W}_0 + \delta \mathbf{W} = \mathbf{W}_0 + \mathbf{AB}\), with \(\mathbf{A} \in \mathbb{R}^{d \times r}\) and \(\mathbf{B} \in \mathbb{R}^{r \times k}\) being trainable low-rank matrices of rank \(r\), significantly smaller in dimensions than \(d\) and \(k\). We finetune a robust baseline to derive the LoRA parameters $\Delta_i$ for task-specific models for $i$-th task, facilitating LLM instruction learning and parameter alignment. SNI is categorized into 60 types based on task types, while P3 encompasses 36 categories, corresponding to 60 and 36 parameter modules, respectively.

\subsubsection{Hypernetwork Pretraining for Preliminary Generalization}
\label{pretraining}
Previous research \citep{demonstration, hypertuning} has demonstrated that pretraining hypernetworks can substantially improve the model's cross-task generalization capabilities. Adhering to the HINT \citep{hint}, we pretrain the hypernetwork on C4 \citep{t5} before finetuning it on a diverse multi-task prompt dataset. As illustrated in the right segment of \autoref{model}, given an input sequence, we partition it into randomly sized segments \(a\), \(b\), and \(c\), where \(a\) is fed into the hypernetwork, \(b\) into the LLM, and \(c\) is the segment to predict. During this stage, training is conducted by minimizing the cross-entropy loss $\mathcal{L}_{\text{pred}}$, aiming to ensure that the hypernetwork learns to recognize instructions to enhance generalization ability.
\begin{equation}
   \mathcal{L}_{\text{pred}} = \text{log}P_{(\text{LLM} + \text{Hypernetwork}(a))}(c~|~b)
\end{equation}

\subsubsection{Hypernetwork Finetuning for Instruction Learning}
\label{finetuning}
At this stage, TAGI is finetuned on a multi-task prompt dataset, enabling it to learn the generation of optimal parameters from task instructions, thereby ensuring effective generalization to future unseen tasks. Similar to the pretraining phase, task instructions (alongside some few-shot samples) replace \(a\), the main input replaces \(b\), and the target replaces \(c\). In each iteration, the hypernetwork generates LoRA parameters and encodes the instructions. LoRA is a parameter-efficient module (i.e., inserting into the model), and the encoded instructions are integrated with the encoder's embeddings for information fusion and concatenated with the fused encoding input during decoding. Beyond the standard $\mathcal{L}_{\text{pred}}$, we employ knowledge distillation for instruction learning: a strong baseline combining complete task instructions and input, serves as the teacher, while the model incorporating generated LoRA parameters with the input, acts as the student. The KL divergence $\mathcal{L}_{\text{kl}}$ measures the discrepancy in word probability distributions between the two models as an implicit learning outcome, and the MSE loss $\mathcal{L}_{\text{ins}}$ calculates the difference between the generated parameters and those of task-specific parameter-efficient modules as an explicit learning intermediate result. The formulation of finetuning is as follows:
\begin{equation}
   \mathcal{L}_{\text{ins}} = \text{MSE}(\Delta_{i}, \text{Hypernetwork}(a))
\end{equation}
\begin{equation}
   \mathcal{L}_{\text{kl}} = \text{KL}(P_{(\text{LLM} + \Delta_{i})}(x~|~(a;b)) ~||~ P_{(\text{LLM} + \text{Hypernetwork}(a))}(x~|~b))
\end{equation}
\begin{equation}
   \mathcal{L}_{\text{finetune}} = \mathcal{L}_{\text{pred}} + \lambda_1 \mathcal{L}_{\text{kl}} + \lambda_2 \mathcal{L}_{\text{ins}}
\end{equation}

where $a \in \mathcal{T}_i$, $\Delta_{i}$ is the optimal LoRA modules of the $i$-th task,  $\lambda_1$ and $\lambda_2$ are the hyper-parameter to control the importance of distillation in finetuning.

\section{Experiments}
\label{main_ex}
We first present the datasets (\cref{data}) and baselines (\cref{base}) used in our evaluation and then discuss three research questions (RQs):

\noindent RQ1: Can the proposed instruction learning paradigm effectively learn the ability of instance training? Can it support cross-task generalization of LLMs? (\cref{main})

\noindent RQ2: How many foundation tasks does TAGI need to learn to achieve better results?  (\cref{task-number})

\noindent RQ3: What is the impact of different modules and learning stages on TAGI? (\cref{ab})

\subsection{Datasets}
\label{data}
To demonstrate the generality of our method, we evaluate our approach on two popular multi-task instruction datasets\footnote{We provide the full list of datasets and more details in the \ref{sec:datasets}.}: \textbf{Super-Natural Instructions (SNI)} \citep{super} and \textbf{T0 split of P3 (P3)} \citep{multitask}. 

% \vspace{0.2cm}
\noindent \textbf{SNI} comprising over 1,600 task datasets, each dataset includes a task definition and a set of fixed positive and negative demonstrations. We follow the previous research \citep{hint, hypertuning} and examine two settings: only using the task definition as the input to the hypernetwork (\textbf{`Def'}), and using the definition along with two few-shot positive examples (\textbf{`Def + 2 Pos'}). We only use the English tasks in the dataset and the model's generation is evaluated on a set of 119 unseen tasks using ROUGE-L.

% \vspace{0.2cm}
\noindent \textbf{P3} composed of 62 task datasets, the T0 model is trained with these tasks divided into meta-training and meta-test sets. The format of the prompts takes into consideration 0-shot reasoning and typically includes instructions or possible answer options. We follow the precedent work \citep{ye-etal-2023-fid} by using the T0 training subset 36 tasks to train our model. The evaluation is conducted based on the accuracy scores of multiple-choice questions for unseen 11 tasks in the meta-test set (MTest11).

\begin{wraptable}[12]{r}{0.6\textwidth}
\vspace{-0.65cm}
  \caption{Compare the characteristics of all comparison methods and the proposed TAGI. More comparisons can be seen in \ref{sec:c_full}.}
  \label{c_half}
  \centering
  \renewcommand\arraystretch{1.05}
  \resizebox{\linewidth}{!}{
      \begin{tabular}{lccccc}
        \toprule
                  & Pre-  & Instr.  & Low Infer. & Instr.    & Unseen  \\
        Method    & Train & Fus.    & Cost       & Learning  & Task \\
        \midrule
        Simple FT   & \XSolidBrush  &  \CheckmarkBold  & \XSolidBrush  & \XSolidBrush & \XSolidBrush \\
        T0 \citep{multitask} / Tk-Instruct \citep{super}   & \XSolidBrush  & \CheckmarkBold  & \XSolidBrush & \XSolidBrush & \CheckmarkBold\CheckmarkBold\CheckmarkBold   \\
        Hypter \citep{ye2021learning}    & \XSolidBrush  & \XSolidBrush  &  \CheckmarkBold & \XSolidBrush  & \CheckmarkBold  \\
        HyperTuning \citep{hypertuning}  & \CheckmarkBold  & \XSolidBrush  &  \CheckmarkBold & \XSolidBrush  & \CheckmarkBold \\
        HINT \citep{hint} & \CheckmarkBold  & \XSolidBrush &  \CheckmarkBold & \XSolidBrush  &  \CheckmarkBold\CheckmarkBold  \\
        \textbf{TAGI (Ours)}    & \CheckmarkBold  & \CheckmarkBold  & \CheckmarkBold & \CheckmarkBold   & \CheckmarkBold\CheckmarkBold\CheckmarkBold \\
        \bottomrule
      \end{tabular}
      }
\end{wraptable}

\subsection{Baselines}
\label{base}
We compare the characteristics of \textbf{TAGI} against eight primary groups of baselines (as shown in \autoref{c_half}): \textit{1)} \textbf{No FT}: models without finetuning. \textit{2)} \textbf{HyperTuning} \citep{hypertuning}: models that use hypernetwork to convert demonstrations into adapters without instruction fusion. \textit{3)} \textbf{Hypter} \citep{ye2021learning}: models based on hypernetwork do not use pretraining. \textit{4)} \textbf{HINT} \citep{hint}: models pretrain hypernetwork and concat instruction. \textit{5)} \textbf{T0} and \textbf{Tk-Instruct}: strong baselines fully finetuned on P3 and SNI respectively with instruction concatenated. \textit{6)} \textbf{Full FT}: models fineuned on target tasks. \textit{7)} \textbf{Decoder-only model}: decoder-only models fully finetuned like GPT-2 \citep{gpt2} and OPT \citep{opt}. \textit{8)} \textbf{FiD-ICL} \citep{ye-etal-2023-fid}: ICL method use encoder intermediate fusion.

\subsection{Implementations}
\label{ex_im}
We limit our scope to encoder-decoder models for our experiments\footnote{We have discussed in detail the encoder-decoder and decoder-only models in \ref{sec:encvsdec}.}. We use T5-LM-Adapt\footnote{\url{https://huggingface.co/google/t5-xl-lm-adapt}} and T0 \citep{multitask} as initializations in our experiments. The two model groups have the same architectural framework but differ in weight; T0 uses T5-LM-Adapt for initialization and undergoes multi-task training on the P3 meta-training set. For \textbf{SNI}, only T5-LM-Adapt is considered, and three different sizes are tested: Base (250M), XL (3B), and XXL (11B), with the teacher model being TK-Instruct \citep{super}. For \textbf{P3}, we experimented with two sets of models of three different sizes: Base (250M), Large (800M), and XL (3B) with the only template as input, while the teacher model being FiD-ICL \citep{ye-etal-2023-fid} with 16-shot examples. The \ref{sec:imp} contains more implementation details and experimental settings.

\subsection{Main Results}
\label{main}

\noindent \textbf{Super-Natural Instructions.} We report the performance and inference costs of TAGI models and baselines in \autoref{m_table}. Our analysis and findings yield several key insights:

$\bullet$ Firstly, \textbf{methods lacking finetuning exhibit subpar performance}. As shown in the first row of the table, the performance of No FT is significantly lower than other baseline methods by approximately 30 points (except for Hypter), which underscores the critical role of inductive bias, introduced during meta-training, in enhancing the model's instructional adherence and cross-task generalization. 

\begin{table}[t]
  \caption{RougeL results on Super-Natural Instructions. The best results are in bold, while the second-best are underlined. $*$, $\dagger$ means that those results are from HINT \citep{hint} and \citep{hypertuning} respectively, "-" means not reported. $\ddagger$ indicates that there is no parameter alignment loss in the hypernetwork finetuning because the model is too large, leading to a significant amount of time required for LoRA tuning for each task. The Average Relative FLOPs cost is calculated relative to Tk-Instruct. We use the number of FLOPs required by each model to process one task (containing 100 examples).}
  \label{m_table}
  \centering
  \renewcommand\arraystretch{1.05}
  \resizebox{\linewidth}{!}{
      \begin{tabular}{lccccccc}
        \toprule
        & \multicolumn{3}{c}{\textbf{Def} (Zero-shot)} & \multicolumn{3}{c}{\textbf{Def + 2 Pos.} (Few-shot)} &  Avg. Rel.  \\
        \cmidrule(r){2-4}\cmidrule(r){5-7}
        \textbf{Method}     & Base (250M) & XL (3B) & XXL (11B) & Base (250M) & XL (3B) & XXL (11B) & FLOPs \\
        \midrule
        No FT & 8.8  & 14.3 & 26.2 & 9.4 & 13.6 & 30.5 & $\times$1.0   \\
        % Full FT$^\dagger$ & 32.4  & 46.6 & - & 41.2 & \underline{54.3} & - & $\times$ 1.0    \\
        Tk-Instruct$^\dagger$   & \textbf{35.3} & \underline{48.0}  & \textbf{53.6} & \underline{42.1} & 54.0 & \textbf{62.0} & $\times$1.0    \\ \hdashline
        \rowcolor{gray!20} \multicolumn{8}{l}{\textit{\# Decoder-only model}}\\
        GPT-2 XL (1.5B)$^*$ & -  & 38.2 & - & - & 45.3 & - & $\times$\textbf{0.33} \\
        OPT (13B)$^*$ & -  & - & 44.8 & - & - & 51.5 & $\times$0.36  \\ \hdashline
        \rowcolor{gray!20} \multicolumn{8}{l}{\textit{\# Hypernetwork-based model}}\\
        Hypter$^*$   & 12.1 & 16.8  & 15.5 &  10.6 &  14.2 & 13.4 & $\times$0.35 \\
        HyperTuning$^\dagger$  & - & 38.9  & - &  - &  48.6 & - & $\times$\underline{0.34} \\
        HINT$^*$  & \underline{33.3} & 47.2  & 51.1 &  41.8 &  53.2 & 56.4 & $\times$0.37  \\
        \textbf{TAGI (Ours)}   & \textbf{35.3}   & \textbf{48.4} & \underline{52.3} $^\ddagger$ & \textbf{42.5} & \textbf{56.3} & \underline{58.4} $^\ddagger$ & $\times$0.39 \\
        \bottomrule
      \end{tabular}
      }
      \vspace{-0.5cm}
\end{table}

$\bullet$ Secondly, \textbf{TAGI demonstrates notable improvements over other hypernetwork-based baselines, with only a marginal increase in inference overhead} (see \autoref{m_table} last column). We find that TAGI still outperforms the advanced method HINT ($\geq 2$ points) while achieving similar computational savings. This highlights the efficacy of instruction learning with knowledge distillation. 
% Such integration facilitates the extraction of supervisory signals from task instructions, aiding the hypernetwork in the automatic generation of more optimal model parameters. 
The underperformance of HINT and Hypertuning may stem from their sole reliance on cross-entropy with the target during meta-training, lacking explicit supervision of intermediate task-specific module parameters and implicit supervision of the teacher outcome. This deficiency impedes their ability to fully leverage instruction tasks for generating superior adapter parameters during meta-test.

$\bullet$ Thirdly, \textbf{TAGI consistently matches or even surpasses robust baselines in both zero- and few-shot settings}. Comparing TAGI with multi-task finetuning approaches such as Full FT and TK-Instruct, we observe that TAGI achieves comparable performance ($0-2.3$ points) except for 11B while utilizing approximately 2.5 $\times$ fewer FLOPs. TAGI's performance on the 11B model is somewhat lacking, potentially attributable to either insufficient training due to resource limitations or a decrement in performance stemming from the omission of parameter alignment constraints due to time constraints\footnote{We discuss the trend and possible reasons in \ref{sec:trend}}. In alignment with prior research, TAGI significantly surpasses GPT-2 and OPT-13B in comparative analyses with decoder-only models ($\geq 10$ points in GPT2 and $\geq 7$ points in OPT-13B), affirming the superiority of encoder-decoder models within similar meta-learning frameworks. Overall, TAGI fulfills its objective by enhancing cross-task generalization capabilities through instruction learning and striking an optimal balance between performance and efficiency.

\begin{table}[t]
  \caption{Average accuracy results over T0 evaluation tasks after training on the T0 P3 train set. $\alpha$ means results are from \citep{ye-etal-2023-fid}. $^\heartsuit$ trained by us followed the Tk-Instruct (meta-training) \citep{super}. Our method uses only template inputs without demonstrations yet achieves competitive performance with ICL-based methods using 16 shots, with much-reduced inference overhead. The Average Relative Inference Time is calculated relative to the Metatrain. We use
  the inference time required by each model to process all 11 test tasks with batch\_size of 1.}
  \label{p3_t}
  \centering
  \renewcommand\arraystretch{1.05}
  \resizebox{\linewidth}{!}{
      \begin{tabular}{lccccccc}
        \toprule
        & \multicolumn{3}{c}{\textbf{T5-LM}} & \multicolumn{3}{c}{\textbf{T0}} & Avg. Rel.  \\
        \cmidrule(r){2-4}\cmidrule(r){5-7}
        \textbf{Method}    & Base (250M) & Large (800M) & XL (3B) & Base (250M) & Large (800M) & XL (3B) & Infer. Time \\
        \midrule
        \rowcolor{blue!5} \multicolumn{8}{l}{\textit{\# MTest11 Avg.}}\\
        Zero-shot & 43.9  & 41.5 & 42.6 & 49.1 & 52.4 & 57.6  & $\times$1.0 \\
        Full FT & 44.6  & 45.5 & 47.2 & \textbf{51.9} & \textbf{56.6} & \textbf{61.4} & $\times$1.0  \\
        Metatrain $^\heartsuit$ & 44.1  & 52.4 & 53.1 & 50.1 & 52.4 & 56.8 & $\times$1.0  \\
         \hdashline
        \rowcolor{gray!20} \multicolumn{8}{l}{\textit{\# ICL-based method}}\\
        Concat-ICL$^{\alpha}$ & 44.2  & 47.6 & - & 48.6 & 53.2 & - & $\times$4.1 \\
        FiD-ICL$^{\alpha}$ & \textbf{47.0}  & \textbf{55.2} & \textbf{60.0} & \underline{51.0} & 53.4 & 58.2 & $\times$1.9  \\ 
        Ensemble-ICL$^{\alpha}$  & 44.6  & 54.5 & 52.6 & 49.9 & 53.7 & 57.7 & $\times$13.2 \\ 
        \hdashline
        \rowcolor{gray!20} \multicolumn{8}{l}{\textit{\# Hypernetwork-based model}}\\
        Hypter$^*$   & - & -  & - &  - &  - & 56.2 & -  \\
        HINT$^*$  & - & -  & - &  - &  - & 60.3 & - \\
        \textbf{TAGI (Ours)}   & \underline{45.6}   & \underline{54.7} & \underline{58.9} & 50.8 & \underline{53.8} & 58.8 & \textbf{$\times$0.88} \\
        \hline
        \rowcolor{blue!5} \multicolumn{8}{l}{\textit{\# HyperT5 Avg. (Without SCloze dataset)}}\\
        FiD-ICL$^{\alpha}$ & \textbf{46.9}  & \textbf{55.8} & 60.6 & \textbf{51.7} & \underline{53.9} & 58.5  & $\times$1.9  \\ 
        HyperTuning$^\dagger$  & - & 54.6 & 59.6 &  - &  - & - & -  \\
        \textbf{TAGI (Ours)}   & \underline{46.7}   & \underline{56.0} & \underline{59.8} & \textbf{51.7} & \textbf{54.6} & \textbf{59.2} & \textbf{$\times$0.88} \\
        \bottomrule
      \end{tabular}
      }
\end{table}

\noindent \textbf{P3.} We report results on the T0 evaluation set in \autoref{p3_t}, with full results in \ref{sec:exres}. 

$\bullet$ Firstly, examining the ICL-based methods presented in the middle section, it is evident that all three ICL aggregation strategies achieve superior performance. This underscores the utility of instructions and demonstrations in aiding LLMs. However, these methods require concatenating extensive demonstrations during both training and inference, which significantly increases computational demands and reduces efficiency ($\times$2 - $\times$13.2 inference time). In contrast, \textbf{TAGI by leveraging solely task instructions one time, attains comparable or superior accuracy levels while significantly curtailing computational burdens ($\times$0.88)}. 
% Comparing TAGI with the ICL-based methods and other baselines, 
TAGI demonstrates a slight disadvantage (merely $1.2$ points) to FiD-ICL \citep{ye-etal-2023-fid} on T5-LM, yet it outperforms other methods ($\geq 1$ point). For T0, it is only 1.5 points lower than Full FT and exceeds all ICL-based methods. Notably, TAGI does not require the 16 examples like the ICL-based method, nor does it necessitate repeated processing of instructions like the baselines, significantly reducing inference overhead.

$\bullet$ A comparison of the first three lines of results indicates that for large or XL models, \textbf{initializing with T5-LM outperforms T0}. We hypothesize that the process of training T5-LM to transition into T0 might result in the dilution of world knowledge or the diminishment of certain specific capabilities, thereby attenuating the benefits derived from meta-training. Conversely, for models of base size, T0 serves as a more effective initialization point.

$\bullet$ Furthermore, \textbf{TAGI outperforms competing hypernetwork models\footnote{ Because HINT is designed for TPU and Hypertuning is not open-sourced, we didn't calculate their inference time. However, based on SNI experiments, it can be inferred that the trend of time expenditure is consistent.}}. 
By comparing the last two columns, it is evident that the performance in MTest11 surpasses HINT and Hypertuning by \(0.5\) and \(4.6\) points respectively. Additionally, in the HyperT5 evaluation, the performance exceeds Hypertuning by \(1\) point. This aligns with prior findings, suggesting that instruction learning augments the hypernetwork's task comprehension and its capacity to generate task-specific adapters.

\subsection{Varying Number of Meta-Training Tasks}
\label{task-number}

A fundamental component of our methodology is incorporating parameter alignment in instruction learning. Consequently, it is imperative to examine the effect of varying the number of tasks on which parameter alignment is applied on outcomes and its influence on the generalization capabilities of LLMs. To this end, we conduct a comprehensive experimental analysis to compare the efficacy of instruction learning with parameter alignment across a spectrum of task quantities against instruction learning devoid of parameter alignment.
Tasks are organized in descending order based on the number of datasets encompassed within each. Subsequently, a predetermined number of tasks are sequentially selected for meta-training purposes. This approach allows us to systematically evaluate the impact of parameter alignment on learning and generalization as the number of tasks varied.

\begin{figure}[t]
\centerline{\includegraphics[width=1.0\textwidth]{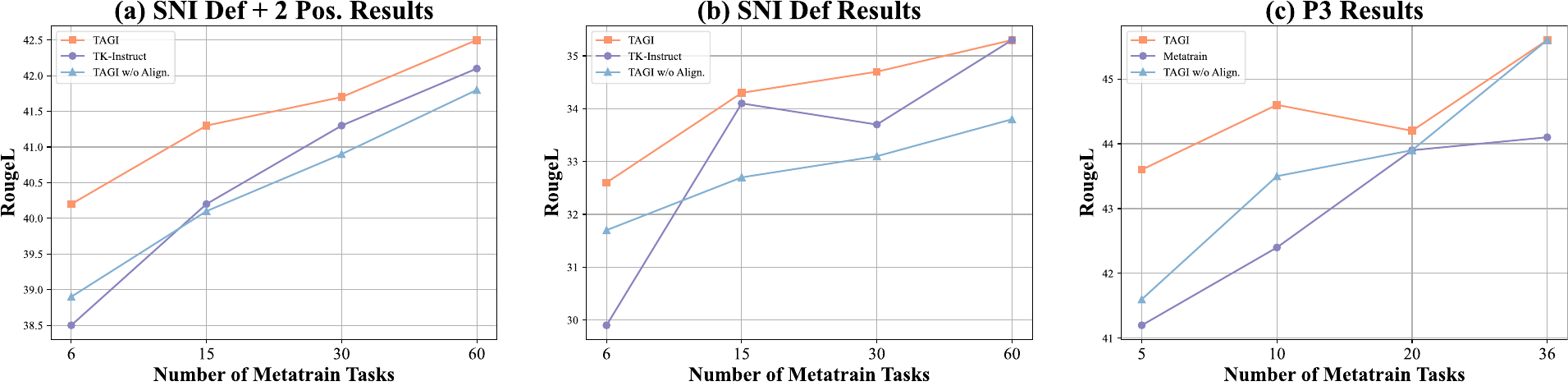}}
% \vspace{-0.45cm}
\caption{The performance of different numbers of meta-training tasks. The backbone model is T5-LM-Base, all trained for 20,000 steps.
}
\label{shot}
% \vspace{-0.65cm}
\end{figure}

From \autoref{shot}, we find that, firstly, \textbf{an increase in the number of tasks correlates with improved performance across all methods}, suggesting that meta-training across a broader array of tasks enhances the model's instruction-following capabilities. However, the practical limitations of sourcing a sufficient quantity of tasks for meta-training must be acknowledged. Secondly, it was observed that the TAGI model exhibits lower overall performance in the absence of parameter alignment for instruction learning, yet it demonstrates a smaller relative standard deviation and less variability in performance in response to the number of tasks. This pattern aligns with the expected outcomes of instruction learning, highlighting the \textbf{efficacy of our approach in bolstering the model's ability to adhere to task instructions and generate task-specific adapters}.

\subsection{Parameter Size against Performance}

\begin{figure}[t]
\centerline{\includegraphics[width=1.0\textwidth]{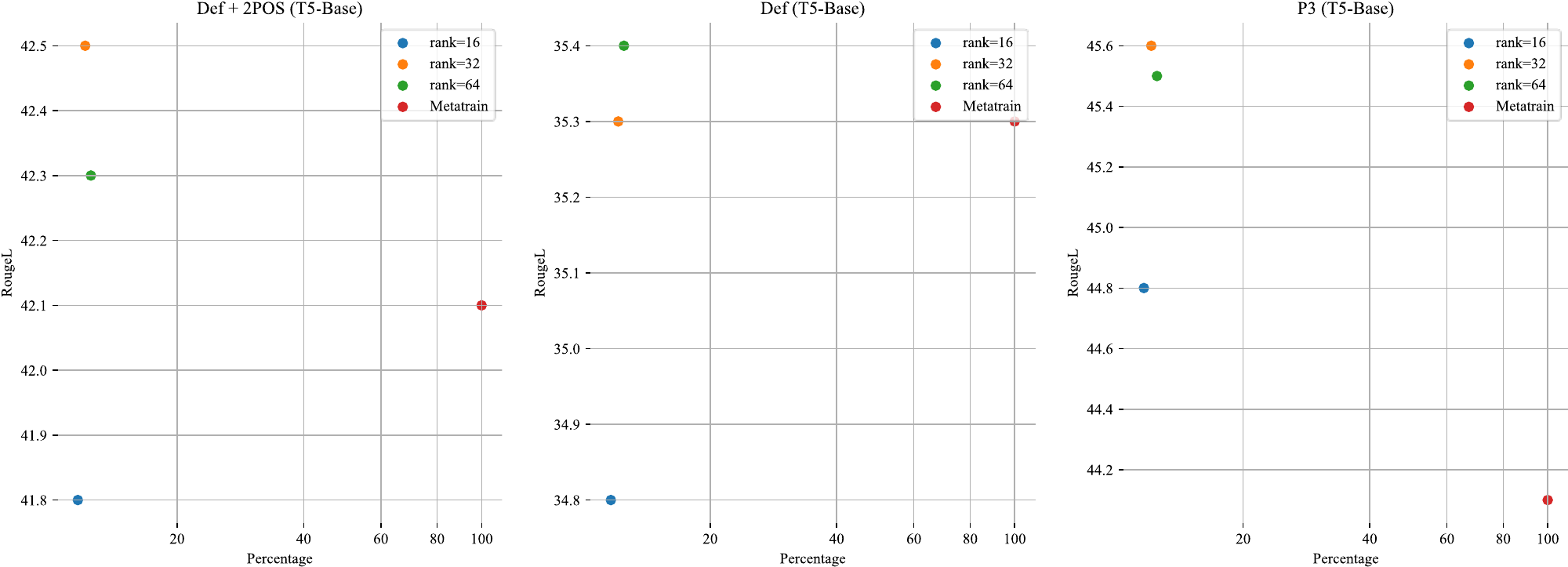}}
% \vspace{-0.45cm}
\caption{The percentage of generated parameters (\%) against performance (RougeL). The backbone model is T5-LM-Base, all trained for 20,000 steps.
}
\label{size}
% \vspace{-0.65cm}
\end{figure}

We analyzed the proportion of generated parameter sizes relative to the total parameter size during the generation of various ranks, and compared this to the performance of the full meta-training fine-tuning method, as demonstrated in Figure \ref{size} and Table \ref{analysis_hyper}. We can find that \textbf{TAGI requires only about 10\% of the parameters to outperform full meta-training fine-tuning which indicates that the limited parameters generated by the Hypernetwork serve as an optimal solution for task completion}. The ability to adaptively construct models tailored to specific tasks removes the necessity for additional fine-tuning, underscoring TAGI's effectiveness and efficiency.

\begin{wraptable}[17]{r}{6cm}
\vspace{-0.65cm}
    \centering
    \caption{Ablation study of TAGI model. All models utilized are T5-LM-XL (3B) and training for 20,000 steps. The P3 dataset was selected by the HyperT5 evaluation.
}
  \label{abl}
  \centering
  \renewcommand\arraystretch{1.05}
  \resizebox{\linewidth}{!}{
      \begin{tabular}{lccc}
        \toprule
        \textbf{Method} & \textbf{Def} & \textbf{Def + 2Pos.} & \textbf{P3} \\
        \midrule
        TK-Instruct  & 48.0 & 54.0 & - \\
        TK-Instruct-LoRA  & 47.5 & 54.6 & - \\
        TK-Instruct-Prefix  & 42.6 & 54.2 & - \\
        \hdashline
        Hypertuning  & 38.9 & 48.6 & 59.6 \\
        HINT  & 47.2 & 53.2 & 60.3 \\
        TAGI  & \textbf{48.4} & \textbf{56.3} & \textbf{60.6} \\
        \hline
        \rowcolor{gray!20} \multicolumn{4}{l}{Ablation Study}\\
        w/o pretraining  & 47.1 & 55.6 & 58.3 \\
        w/o Instr. Fus.  & 35.1 & 40.6 & 44.2 \\
        w/o $\mathcal{L}_{\text{ce}}$   & 47.6 & 55.4 & 59.8 \\
        w/o $\mathcal{L}_{\text{kl}}$   & 45.7 & 53.9 & 57.3 \\
        w/o $\mathcal{L}_{\text{ins}}$  & 47.5 & 55.2 & 59.4 \\
        w/o Hypernetwork & 43.8 & 50.7 & -\\
        \bottomrule
      \end{tabular}
      }
\end{wraptable}

\subsection{Ablation Study}
\label{ab}
To evaluate the significance of each component within the TAGI model, we conducted a series of experiments across two meta-task datasets utilizing the T5-LM-XL (3B) model. The results as depicted in the \autoref{abl}, highlight that the \textbf{instructions fusion plays a pivotal role in enhancing model performance}. This process facilitates dynamic interaction between the input and the instructions, enriching the model's input with additional contextual information, reminiscent of the substantial benefits observed with ICL. Moreover, \textbf{pretraining emerges as a critical phase}, markedly improving the capabilities of models that have not undergone pretraining, thereby significantly enhancing their proficiency in interpreting and executing task instructions. Furthermore, the systematic removal of \textbf{various components during the finetuning phase indicates a consistent decline in performance}, underscoring the integral contribution of each component to the model's overall efficacy.

Compared to meta-learning methods such as LoRA fine-tuning (rank=32) "Tk-Instruct-LoRA", prefix fine-tuning (num\_virtual\_tokens=32) "Tk-Instruct-prefix", and full fine-tuning "Tk-Instruct", our TAGI method enhances task comprehension and utilization which achieved through a hypernetwork that dynamically generates adapter LoRA insertions into the LLM based on input, leads to better cross-task generalization capabilities. Notably, prefix fine-tuning excels in the Def + 2Pos scenario, likely due to its effective integration of information from positive examples. Conversely, the Def scenario performs less satisfactorily, indicating that instructions alone are insufficient for optimal results. Comparative analysis with other hypernetwork models reveals that TAGI's ablation performance remains robust, affirming the effectiveness of each step in bolstering TAGI's operational efficiency.

\section{Conclusions}
In this paper, we introduce an innovative method of instruction learning designed to emulate instance training. This approach enables the model to achieve specified tasks and learn from instructions on how to address a category of problems. The proposed TAGI seamlessly integrates instruction into the input and processes the instruction simultaneously, thereby ensuring minimal inference overhead. Concurrently, we employ a knowledge distillation framework to facilitate instruction learning for distilling skills and aligning task-specific models. This allows the hypernetwork to transform task instructions into an efficient module inserted into the LLMs, thereby boosting generalization performance. Remarkably, TAGI consistently equals or surpasses the efficacy of conventional meta-training approaches while requiring fewer FLOPs and obviating the need for additional model parameters updating or gradient back-propagation. Future work will investigate more potent hypernetwork pretraining techniques and develop superior instruction fusion methods to augment the hypernetwork's expressive capability, thereby enhancing the model's ability to generalize to unseen tasks. Moreover, future work will investigate various task type classifications and the generalization effects of cross-modal tasks in instruction learning.

\section{Acknowledgements}

This work was supported by National Key R\&D Program of China (No. 2022YFF0711900) and the National Natural Science Foundation of China (No.62376270, No.62276264). This work was supported by the Youth Innovation Promotion Association CAS.

\bibliography{custom}

\begin{thebibliography}{10}

\bibitem{meta-learn}
Jonathan Baxter.
\newblock {\em Learning to Learn}.
\newblock Springer US, 1998.

\bibitem{multilingual}
Christos Baziotis, Mikel Artetxe, James Cross, and Shruti Bhosale.
\newblock Multilingual machine translation with hyper-adapters, 2022.

\bibitem{gpt3}
Tom Brown, Benjamin Mann, Nick Ryder, Melanie Subbiah, Jared~D Kaplan, Prafulla Dhariwal, Arvind Neelakantan, Pranav Shyam, Girish Sastry, Amanda Askell, Sandhini Agarwal, Ariel Herbert-Voss, Gretchen Krueger, Tom Henighan, Rewon Child, Aditya Ramesh, Daniel Ziegler, Jeffrey Wu, Clemens Winter, Chris Hesse, Mark Chen, Eric Sigler, Mateusz Litwin, Scott Gray, Benjamin Chess, Jack Clark, Christopher Berner, Sam McCandlish, Alec Radford, Ilya Sutskever, and Dario Amodei.
\newblock Language models are few-shot learners.
\newblock In H.~Larochelle, M.~Ranzato, R.~Hadsell, M.F. Balcan, and H.~Lin, editors, {\em Advances in Neural Information Processing Systems}, volume~33, pages 1877--1901. Curran Associates, Inc., 2020.

\bibitem{brief}
Vinod~Kumar Chauhan, Jiandong Zhou, Ping Lu, Soheila Molaei, and David~A. Clifton.
\newblock A brief review of hypernetworks in deep learning, 2023.

\bibitem{demonstration}
Tong Chen, Qirun Dai, Zhijie Deng, and Dequan Wang.
\newblock Demonstration distillation for efficient in-context learning, 2024.

\bibitem{scaling}
Hyung~Won Chung, Le~Hou, Shayne Longpre, Barret Zoph, Yi~Tay, William Fedus, Yunxuan Li, Xuezhi Wang, Mostafa Dehghani, Siddhartha Brahma, Albert Webson, Shixiang~Shane Gu, Zhuyun Dai, Mirac Suzgun, Xinyun Chen, Aakanksha Chowdhery, Alex Castro-Ros, Marie Pellat, Kevin Robinson, Dasha Valter, Sharan Narang, Gaurav Mishra, Adams Yu, Vincent Zhao, Yanping Huang, Andrew Dai, Hongkun Yu, Slav Petrov, Ed~H. Chi, Jeff Dean, Jacob Devlin, Adam Roberts, Denny Zhou, Quoc~V. Le, and Jason Wei.
\newblock Scaling instruction-finetuned language models, 2022.

\bibitem{deb2022boosting}
Budhaditya Deb, Guoqing Zheng, and Ahmed~Hassan Awadallah.
\newblock Boosting natural language generation from instructions with meta-learning, 2022.

\bibitem{hypernetworks}
David Ha, Andrew Dai, and Quoc~V. Le.
\newblock Hypernetworks, 2016.

\bibitem{hyperprompt}
Yun He, Huaixiu~Steven Zheng, Yi~Tay, Jai Gupta, Yu~Du, Vamsi Aribandi, Zhe Zhao, YaGuang Li, Zhao Chen, Donald Metzler, Heng-Tze Cheng, and Ed~H. Chi.
\newblock Hyperprompt: Prompt-based task-conditioning of transformers, 2022.

\bibitem{kd}
Geoffrey Hinton, Oriol Vinyals, and Jeff Dean.
\newblock Distilling the knowledge in a neural network.
\newblock {\em arXiv preprint arXiv:1503.02531}, 2015.

\bibitem{houlsby2019parameterefficient}
Neil Houlsby, Andrei Giurgiu, Stanislaw Jastrzebski, Bruna Morrone, Quentin de~Laroussilhe, Andrea Gesmundo, Mona Attariyan, and Sylvain Gelly.
\newblock Parameter-efficient transfer learning for nlp, 2019.

\bibitem{hu2022lora}
Edward~J Hu, yelong shen, Phillip Wallis, Zeyuan Allen-Zhu, Yuanzhi Li, Shean Wang, Lu~Wang, and Weizhu Chen.
\newblock Lo{RA}: Low-rank adaptation of large language models.
\newblock In {\em International Conference on Learning Representations}, 2022.

\bibitem{hint}
Hamish Ivison, Akshita Bhagia, Yizhong Wang, Hannaneh Hajishirzi, and Matthew Peters.
\newblock Hint: Hypernetwork instruction tuning for efficient zero-shot generalisation.
\newblock {\em ACL}, 2023.

\bibitem{KL}
James~M. Joyce.
\newblock {\em Kullback-Leibler Divergence}, pages 720--722.
\newblock Springer Berlin Heidelberg, Berlin, Heidelberg, 2011.

\bibitem{teach}
Sharon Kim, Mahjabeen Raza, and Edward Seidman.
\newblock {\em Improving 21st-century teaching skills: The key to effective 21st-century learners}.
\newblock Springer US, 2019.

\bibitem{datasets}
Quentin Lhoest, Albert Villanova~del Moral, Yacine Jernite, Abhishek Thakur, Patrick von Platen, Suraj Patil, Julien Chaumond, Mariama Drame, Julien Plu, Lewis Tunstall, Joe Davison, Mario {\v{S}}a{\v{s}}ko, Gunjan Chhablani, Bhavitvya Malik, Simon Brandeis, Teven Le~Scao, Victor Sanh, Canwen Xu, Nicolas Patry, Angelina McMillan-Major, Philipp Schmid, Sylvain Gugger, Cl{\'e}ment Delangue, Th{\'e}o Matussi{\`e}re, Lysandre Debut, Stas Bekman, Pierric Cistac, Thibault Goehringer, Victor Mustar, Fran{\c{c}}ois Lagunas, Alexander Rush, and Thomas Wolf.
\newblock Datasets: A community library for natural language processing.
\newblock In Heike Adel and Shuming Shi, editors, {\em Proceedings of the 2021 Conference on Empirical Methods in Natural Language Processing: System Demonstrations}, pages 175--184, Online and Punta Cana, Dominican Republic, November 2021. Association for Computational Linguistics.

\bibitem{prefix}
Xiang~Lisa Li and Percy Liang.
\newblock Prefix-tuning: Optimizing continuous prompts for generation.
\newblock In Chengqing Zong, Fei Xia, Wenjie Li, and Roberto Navigli, editors, {\em Proceedings of the 59th Annual Meeting of the Association for Computational Linguistics and the 11th International Joint Conference on Natural Language Processing (Volume 1: Long Papers)}, pages 4582--4597, Online, August 2021. Association for Computational Linguistics.

\bibitem{liao2025awakening}
Huanxuan Liao, Shizhu He, Yao Xu, Yuanzhe Zhang, Shengping Liu, Kang Liu, and Jun Zhao.
\newblock Awakening augmented generation: Learning to awaken internal knowledge of large language models for question answering.
\newblock In {\em Proceedings of the 31st International Conference on Computational Linguistics}, pages 1333--1352, 2025.

\bibitem{fewshot}
Haokun Liu, Derek Tam, Mohammed Muqeeth, Jay Mohta, Tenghao Huang, Mohit Bansal, and Colin Raffel.
\newblock Few-shot parameter-efficient fine-tuning is better and cheaper than in-context learning, 2022.

\bibitem{longpre2023flan}
Shayne Longpre, Le~Hou, Tu~Vu, Albert Webson, Hyung~Won Chung, Yi~Tay, Denny Zhou, Quoc~V. Le, Barret Zoph, Jason Wei, and Adam Roberts.
\newblock The flan collection: Designing data and methods for effective instruction tuning, 2023.

\bibitem{metaicl}
Sewon Min, Mike Lewis, Luke Zettlemoyer, and Hannaneh Hajishirzi.
\newblock {M}eta{ICL}: Learning to learn in context.
\newblock In Marine Carpuat, Marie-Catherine de~Marneffe, and Ivan~Vladimir Meza~Ruiz, editors, {\em Proceedings of the 2022 Conference of the North American Chapter of the Association for Computational Linguistics: Human Language Technologies}, pages 2791--2809, Seattle, United States, July 2022. Association for Computational Linguistics.

\bibitem{reframing}
Swaroop Mishra, Daniel Khashabi, Chitta Baral, Yejin Choi, and Hannaneh Hajishirzi.
\newblock Reframing instructional prompts to gptk's language, 2022.

\bibitem{naturalinstructions}
Swaroop Mishra, Daniel Khashabi, Chitta Baral, and Hannaneh Hajishirzi.
\newblock Cross-task generalization via natural language crowdsourcing instructions.
\newblock In {\em ACL}, 2022.

\bibitem{crosstask}
Swaroop Mishra, Daniel Khashabi, Chitta Baral, and Hannaneh Hajishirzi.
\newblock Cross-task generalization via natural language crowdsourcing instructions, 2022.

\bibitem{training}
Long Ouyang, Jeff Wu, Xu~Jiang, Diogo Almeida, Carroll~L. Wainwright, Pamela Mishkin, Chong Zhang, Sandhini Agarwal, Katarina Slama, Alex Ray, John Schulman, Jacob Hilton, Fraser Kelton, Luke Miller, Maddie Simens, Amanda Askell, Peter Welinder, Paul Christiano, Jan Leike, and Ryan Lowe.
\newblock Training language models to follow instructions with human feedback, 2022.

\bibitem{pytorch}
Adam Paszke, Sam Gross, Francisco Massa, Adam Lerer, James Bradbury, Gregory Chanan, Trevor Killeen, Zeming Lin, Natalia Gimelshein, Luca Antiga, Alban Desmaison, Andreas K\"{o}pf, Edward Yang, Zach DeVito, Martin Raison, Alykhan Tejani, Sasank Chilamkurthy, Benoit Steiner, Lu~Fang, Junjie Bai, and Soumith Chintala.
\newblock {\em PyTorch: an imperative style, high-performance deep learning library}.
\newblock Curran Associates Inc., Red Hook, NY, USA, 2019.

\bibitem{hypertuning}
Jason Phang, Yi~Mao, Pengcheng He, and Weizhu Chen.
\newblock Hypertuning: Toward adapting large language models without back-propagation, 2022.

\bibitem{gpt2}
Alec Radford, Jeff Wu, Rewon Child, David Luan, Dario Amodei, and Ilya Sutskever.
\newblock Language models are unsupervised multitask learners.
\newblock 2019.

\bibitem{t5}
Colin Raffel, Noam Shazeer, Adam Roberts, Katherine Lee, Sharan Narang, Michael Matena, Yanqi Zhou, Wei Li, and Peter~J. Liu.
\newblock Exploring the limits of transfer learning with a unified text-to-text transformer.
\newblock {\em Journal of Machine Learning Research}, 21(140):1--67, 2020.

\bibitem{multitask}
Victor Sanh, Albert Webson, Colin Raffel, Stephen~H. Bach, Lintang Sutawika, Zaid Alyafeai, Antoine Chaffin, Arnaud Stiegler, Teven~Le Scao, Arun Raja, Manan Dey, M~Saiful Bari, Canwen Xu, Urmish Thakker, Shanya~Sharma Sharma, Eliza Szczechla, Taewoon Kim, Gunjan Chhablani, Nihal Nayak, Debajyoti Datta, Jonathan Chang, Mike Tian-Jian Jiang, Han Wang, Matteo Manica, Sheng Shen, Zheng~Xin Yong, Harshit Pandey, Rachel Bawden, Thomas Wang, Trishala Neeraj, Jos Rozen, Abheesht Sharma, Andrea Santilli, Thibault Fevry, Jason~Alan Fries, Ryan Teehan, Tali Bers, Stella Biderman, Leo Gao, Thomas Wolf, and Alexander~M. Rush.
\newblock Multitask prompted training enables zero-shot task generalization, 2022.

\bibitem{Schmidhuber1992LearningTC}
J{\"u}rgen Schmidhuber.
\newblock Learning to control fast-weight memories: An alternative to dynamic recurrent networks.
\newblock {\em Neural Computation}, 4:131--139, 1992.

\bibitem{learningbydistill}
Charlie Snell, Dan Klein, and Ruiqi Zhong.
\newblock Learning by distilling context, 2022.

\bibitem{hypergrid}
Yi~Tay, Zhe Zhao, Dara Bahri, Donald Metzler, and Da-Cheng Juan.
\newblock Hypergrid transformers: Towards a single model for multiple tasks.
\newblock In {\em International Conference on Learning Representations}, 2021.

\bibitem{LearningTL}
Sebastian Thrun and Lorien~Y. Pratt.
\newblock Learning to learn: Introduction and overview.
\newblock In {\em Learning to Learn}, 1998.

\bibitem{language}
Thomas Wang, Adam Roberts, Daniel Hesslow, Teven~Le Scao, Hyung~Won Chung, Iz~Beltagy, Julien Launay, and Colin Raffel.
\newblock What language model architecture and pretraining objective work best for zero-shot generalization?, 2022.

\bibitem{minilm}
Wenhui Wang, Furu Wei, Li~Dong, Hangbo Bao, Nan Yang, and Ming Zhou.
\newblock Minilm: Deep self-attention distillation for task-agnostic compression of pre-trained transformers.
\newblock {\em Advances in Neural Information Processing Systems}, 33:5776--5788, 2020.

\bibitem{super}
Yizhong Wang, Swaroop Mishra, Pegah Alipoormolabashi, Yeganeh Kordi, Amirreza Mirzaei, Atharva Naik, Arjun Ashok, Arut~Selvan Dhanasekaran, Anjana Arunkumar, David Stap, Eshaan Pathak, Giannis Karamanolakis, Haizhi Lai, Ishan Purohit, Ishani Mondal, Jacob Anderson, Kirby Kuznia, Krima Doshi, Kuntal~Kumar Pal, Maitreya Patel, Mehrad Moradshahi, Mihir Parmar, Mirali Purohit, Neeraj Varshney, Phani~Rohitha Kaza, Pulkit Verma, Ravsehaj~Singh Puri, Rushang Karia, Savan Doshi, Shailaja~Keyur Sampat, Siddhartha Mishra, Sujan Reddy~A, Sumanta Patro, Tanay Dixit, and Xudong Shen.
\newblock Super-{N}atural{I}nstructions: Generalization via declarative instructions on 1600+ {NLP} tasks.
\newblock In Yoav Goldberg, Zornitsa Kozareva, and Yue Zhang, editors, {\em Proceedings of the 2022 Conference on Empirical Methods in Natural Language Processing}, pages 5085--5109, Abu Dhabi, United Arab Emirates, December 2022. Association for Computational Linguistics.

\bibitem{flan}
Jason Wei, Maarten Bosma, Vincent Zhao, Kelvin Guu, Adams~Wei Yu, Brian Lester, Nan Du, Andrew~M. Dai, and Quoc~V Le.
\newblock Finetuned language models are zero-shot learners.
\newblock In {\em International Conference on Learning Representations}, 2022.

\bibitem{learning}
Orion Weller, Nicholas Lourie, Matt Gardner, and Matthew~E. Peters.
\newblock Learning from task descriptions.
\newblock In Bonnie Webber, Trevor Cohn, Yulan He, and Yang Liu, editors, {\em Proceedings of the 2020 Conference on Empirical Methods in Natural Language Processing (EMNLP)}, pages 1361--1375, Online, November 2020. Association for Computational Linguistics.

\bibitem{transformers}
Thomas Wolf, Lysandre Debut, Victor Sanh, Julien Chaumond, Clement Delangue, Anthony Moi, Pierric Cistac, Tim Rault, Remi Louf, Morgan Funtowicz, Joe Davison, Sam Shleifer, Patrick von Platen, Clara Ma, Yacine Jernite, Julien Plu, Canwen Xu, Teven Le~Scao, Sylvain Gugger, Mariama Drame, Quentin Lhoest, and Alexander Rush.
\newblock Transformers: State-of-the-art natural language processing.
\newblock In Qun Liu and David Schlangen, editors, {\em Proceedings of the 2020 Conference on Empirical Methods in Natural Language Processing: System Demonstrations}, pages 38--45, Online, October 2020. Association for Computational Linguistics.

\bibitem{ye-etal-2023-fid}
Qinyuan Ye, Iz~Beltagy, Matthew Peters, Xiang Ren, and Hannaneh Hajishirzi.
\newblock {F}i{D}-{ICL}: A fusion-in-decoder approach for efficient in-context learning.
\newblock In Anna Rogers, Jordan Boyd-Graber, and Naoaki Okazaki, editors, {\em Proceedings of the 61st Annual Meeting of the Association for Computational Linguistics (Volume 1: Long Papers)}, pages 8158--8185, Toronto, Canada, July 2023. Association for Computational Linguistics.

\bibitem{ye2021learning}
Qinyuan Ye and Xiang Ren.
\newblock Learning to generate task-specific adapters from task description, 2021.

\bibitem{opt}
Susan Zhang, Stephen Roller, Naman Goyal, Mikel Artetxe, Moya Chen, Shuohui Chen, Christopher Dewan, Mona Diab, Xian Li, Xi~Victoria Lin, Todor Mihaylov, Myle Ott, Sam Shleifer, Kurt Shuster, Daniel Simig, Punit~Singh Koura, Anjali Sridhar, Tianlu Wang, and Luke Zettlemoyer.
\newblock Opt: Open pre-trained transformer language models, 2022.

\bibitem{adapting}
Ruiqi Zhong, Kristy Lee, Zheng Zhang, and Dan Klein.
\newblock Adapting language models for zero-shot learning by meta-tuning on dataset and prompt collections, 2021.

\bibitem{hyperx}
Ahmet Üstün, Arianna Bisazza, Gosse Bouma, Gertjan van Noord, and Sebastian Ruder.
\newblock Hyper-x: A unified hypernetwork for multi-task multilingual transfer, 2022.

\end{thebibliography}
\bibliographystyle{plain}

% \cite{blub}
% \bibliographystyle{plainnat}
% % \bibliographystyle{unsrtnat}
% \bibliography{custom}

%%%%%%%%%%%%%%%%%%%%%%%%%%%%%%%%%%%%%%%%%%%%%%%%%%%%%%%%%%%%

\appendix

\section{Experimantal Settings}
\label{experiment}
\subsection{Problem Setting}
\noindent \textbf{Meta-Training and Inference:} Our methodology rigorously adheres to the protocol outlined in MetaICL \citep{metaicl}. In the meta-train phase, we commence by selecting a task $\mathcal{T}$ from $\mathcal{T}_{train}$, followed by the sampling of $k$ support examples $\{(x^{(s)}_i, y^{(s)}_i)\}$ and $m$ query examples $\{(x^{(q)}_i, y^{(q)}_i)\}$ from the chosen task. The proposed hypernetwork is then adjusted to minimize the overall loss, focusing on generating a task model that can accurately predict the target sequences (e.g., answer) for source sequences (e.g. question). During the meta-test/inference phase, for each novel task in $\mathcal{T}_{test}$, we employ instructions to create the task-specific adapter, to optimize the model's performance across all query examples $\{(x^{(q)}_i, y^{(q)}_i)\}$.

\begin{table}[htbp]
  \caption{Number of samples in given splits for each dataset.}
  \label{data_d}
  \centering
  \renewcommand\arraystretch{1.05}
  % \resizebox{\linewidth}{!}{
      \begin{tabular}{lccc}
        \toprule
        Dataset & Examples per Task & Train & Test \\
        \midrule
        Super-Natural Instructions & 100 & 75,417 & 11,810 \\
        P3 & - & 90,897,454 & 2,940,068 \\
        P3 (Sampling) & 1000 & 290,000 & 2,940,068 \\
        \bottomrule
      \end{tabular}
      % }
\end{table}

\subsection{Datasets}
\label{sec:datasets}
During the pretraining phase, we utilized the C4 dataset \citep{t5}, truncating each sequence to 1024 tokens. For the training phase, we employed Super-Natural Instructions (SNI) \citep{super} and P3 datasets \citep{multitask} for meta-training and meta-test. For SNI, we adhered to the default settings \citep{hint, super}, which include 100 examples per task for both the training and test splits. For P3, we used the data and prompts provided by T0. All prompts related to the meta-training tasks were included in the meta-training process, while the meta-test phase utilized evaluation prompts specified by T0 \citep{multitask}. We treated ANLI R1, R2, and R3 as three distinct tasks, resulting in 11 tasks for the original meta-test in P3 (Meta-Test-11). Due to resource constraints, we deviated from the sampling procedures of prior work, opting to sample 1000 examples per task for each prompt template. This approach yielded a smaller dataset size, as detailed in \autoref{data_d}. For further information on P3 refer to \citep{multitask}. Additionally, to facilitate comparison with the Hypertuning method, we excluded the StoryCloze task from the evaluation since it was not included in the datasets for the HyperT5 evaluation.

\subsection{Split Sizes for Varying Number of Meta-Training Tasks}
As shown in \autoref{sni_list} and \autoref{p3_list}, we present a comprehensive list of the two datasets, including the number of tasks or templates contained in each task and the task divisions from \cref{task-number} experiments. The divisions in the table are cumulative; thus, the second division includes both the first and the second divisions. For SNI, tasks were sorted in descending order based on the number of tasks they contained and then divided into specified sizes (6, 15, 30, 60). For P3, we selected a specified number of tasks (5, 10, 20, 36) based on the task classification in the original paper, which includes categories such as Multiple-Choice QA, Closed-Book QA, Summarization, Structure-To-Text, Paraphrase Identification, Sentiment, Topic Classification, and Extractive QA.

We obtain all our data from huggingface datasets \citep{datasets}. In the following, we provide the dataset links:
\begin{itemize}
    \item Super-Natural Instructions: \url{https://github.com/allenai/natural-instructions}
    \item P3: \url{https://huggin gface.co/datasets/bigscience/P3}
\end{itemize}

Additionally, the Super-Natural Instructions dataset (previously known as Natural Instructions-v2) has undergone some changes over time. In our experiments, we use the v2.6 version.

\subsection{Implementations}
\label{sec:imp}
Our implementations are based on huggingface transformers v4.23.1 \citep{transformers} using PyTorch v1.13.1 \citep{pytorch} and deepspeed\footnote{\url{https://github.com/microsoft/DeepSpeed}} v0.10.0. All experiments were conducted on 4 A100 NVIDIA GPUs, each equipped with 80GB of memory, and eight A6000  NVIDIA GPUs with 48GB of memory. Unless otherwise specified, the rank of LoRA generated by the hypernetwork is 32, and we use the Adamw optimizer with a learning rate of 5e-5 and a linear warmup rate of 0.02. We pre-train all models for 50,000 steps using C4 \citep{t5} with a batch size of 8 samples and sequences of length 1024.

\subsection{T0-Base/Large/3B}
T0 \citep{multitask} provides model checkpoints only in sizes 3B and 11B. Additionally, HINT \citep{hint} and FiD-ICL \citep{ye-etal-2023-fid} re-pretrained T0 and found that the model was not sufficiently trained, achieving better results after reproduction. Therefore, we used the T0 model \footnote{\url{https://huggingface.co/qinyuany/fid-icl-t0-large}} reproduced by FiD-ICL to conduct a series of experiments. 

\begin{table}[htbp]
  \caption{Hyperparameters for Training TAGI Models and LoRA Tuning.}
  \label{hyperparam}
  \centering
  \renewcommand\arraystretch{1.05}
   \resizebox{\linewidth}{!}{
  % \resizebox{\linewidth}{!}{
      \begin{tabular}{lcccccccc}
        \toprule
          &  & & \multicolumn{6}{c}{\textbf{Finetuning}} \\
          &  & & \multicolumn{3}{c}{\textbf{SNI}} & \multicolumn{3}{c}{\textbf{P3}} \\
           \cmidrule(r){4-6}\cmidrule(r){7-9}
         & \textbf{LoRA Tuning} &  \textbf{Pretraining} & Base (250M) & XL(3B) & XXL (11B) & Base (250M) & Large (800M) & XL(3B)\\
        \midrule
        Max Input Len & 1024 & 1024 & 1024 & 1024 & 1024 & 512 & 512 & 512 \\
        Max Output Len & 128 & - & 128 & 128 & 128 & 64 & 64 & 64  \\
        Optimizer & adamw & adafactor  & adamw & adamw & adamw & adamw & adamw  & adamw \\
        Learning Rate & 1e-4 & 1e-3  & 1e-4 & 5e-5 & 5e-5 & 1e-4 & 1e-4 & 5e-5 \\
        precision & bf16 & float32 & bf16 & bf16 & bf16 & bf16 & bf16 & bf16 \\
        \# Training Steps & 10000 & 50000 & 20000  & 20000 & 20000 & 20000 & 20000 & 20000\\
        \rowcolor{gray!20} \cellcolor{white}\# Warmup Steps & \cellcolor{white}- & \cellcolor{white}- & \multicolumn{6}{c}{\# 2\% of total training steps} \\
        Batch Size & 8 & 8 & 8 & 2 & 1 & 8 & 4 & 2  \\
        Gradient Accumulation & 2 & 1 & 2 & 4 & 2 & 2 & 4 & 4 \\
        \rowcolor{gray!20} \cellcolor{white}LoRA Rank & \multicolumn{8}{c}{\# 32} \\
        \bottomrule
      \end{tabular}
      }
      % }
\end{table}

\subsection{Hyperparameter}
The complete stable hyperparameter set used for training runs can be found in \autoref{hyperparam}.

\section{Additional Experiments and Findings}
\subsection{Why we choose Enc-Dec Models?}
\label{sec:encvsdec}
Previous work has suggested that models with an encoder-decoder (enc-dec) structure have advantages over decoder-only (dec-only) models in terms of task generalization and instruction-following capabilities \citep{longpre2023flan, language, ye-etal-2023-fid}. Therefore, in our experiments, we only considered models with an enc-dec structure (T5-LM and T0). Our experimental results demonstrated that enc-dec models indeed have an advantage when compared, although dec-only models might have higher computational efficiency due to their ability to cache KV and have fewer layers. However, our method, TAGI, significantly improves performance in various aspects with only a slight increase in computational overhead. We encode the task instructions only once based on the original computation.

\subsection{T5-LM-XXL Training Trend}
\label{sec:trend}
\begin{figure}[t]
\centerline{\includegraphics[width=1.0\textwidth]{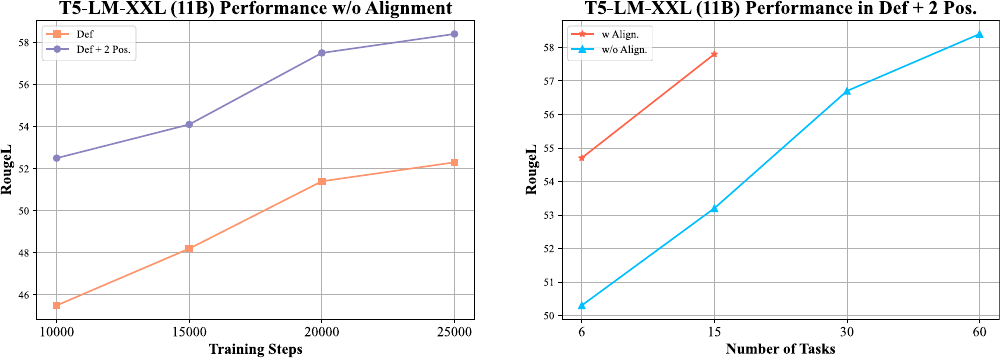}}
% \vspace{-0.45cm}
\caption{Analysis of T5-LM-XXL (11B).
}
\label{trend}
% \vspace{-0.65cm}
\end{figure}

In this section, we detail how the performance of the T5-LM-XXL (11B) model surpasses the hypernetwork models but falls short of the meta-trained strong baseline Tk-Instruct by 1-4 points, as mentioned earlier in \cref{main}. The primary reason is insufficient training; when replicating the Tk-Instruct experiment, our results were significantly lower than reported when finetuning for only 20,000 steps. Consequently, we analyzed the performance of our TAGI model at different finetuning steps. As shown on the left side of \autoref{trend}, performance steadily improves with more steps with substantial growth. Thus, we reasonably predict that increasing the steps to 50,000 or more could surpass Tk-Instruct. Another possible reason is the lack of parameter alignment for the 11B model due to limited resources. Our previous analysis has shown that parameter alignment is crucial, with larger models benefiting more. Therefore, we analyzed performance with a small number of tasks for parameter alignment. As shown on the right side of \autoref{trend}, performance with parameter alignment for 6 and 15 tasks is better than without alignment. Based on these trends, it can be inferred that performance with full task parameter alignment could surpass Tk-Instruct.

\subsection{Analysis on Hyperparameters}
\label{sec:analysis}
To explore the optimal hyperparameter settings for our experiments, we conducted a series of tests and error analyses using the T5-LM-Base (800M) model. The findings presented in \autoref{analysis_hyper} reveal that variations in hyperparameters can lead to performance fluctuations, particularly with higher learning rates or reduced finetuning steps. Given the varying pre-training conditions of models of different sizes, a size-specific analysis is essential; however, details on larger models are omitted here due to resource limitations.

We observed that different settings of LoRA minimally affect performance, leading us to select a balanced size of 32. Similarly, the impact of the warmup ratio is negligible; thus, based on our experience, we chose a warmup ratio of one percent of the maximum finetuning steps. While more finetuning steps generally correlate with improved performance, excessive finetuning can result in overfitting on meta-training tasks, thereby diminishing generalizability. Moreover, increased finetuning steps require greater computational resources. Consequently, we determined that the optimal number of finetuning steps is 20,000 based on our experimental outcomes.
\begin{table}[htbp]
  \caption{Performance variation due to different hyperparameters. The base model is T5-LM-Base, and all experiments follow the previous hyperparameter settings, changing only the target hyperparameter, where underlines indicate experimental defaults.}
  \label{analysis_hyper}
  \centering
  \renewcommand\arraystretch{1.05}
   \resizebox{\linewidth}{!}{
  % \resizebox{\linewidth}{!}{
      \begin{tabular}{lccccccccccccc}
        \toprule
          & \multicolumn{4}{c}{\textbf{Learning Rate}} & \multicolumn{3}{c}{\textbf{LoRA Rank}}  & \multicolumn{3}{c}{\textbf{Training Steps}} & \multicolumn{3}{c}{\textbf{Warmup Ratio}}\\
           \cmidrule(r){2-5}\cmidrule(r){6-8}\cmidrule(r){9-11}\cmidrule(r){12-14}
         \textbf{Method} & 5e-5 &  \underline{1e-4} & 3e-4 & 1e-3 & 16 & \underline{32} & 64 & 15000 & \underline{20000} & 25000 & 0.01 & \underline{0.02} & 0.03\\
        \midrule
        \rowcolor{green!10}\multicolumn{14}{l}{\textit{\textbf{SNI}}} \\ 
        \rowcolor{blue!10}\multicolumn{14}{l}{\textit{Def + 2 Pos.}} \\ \hline
        \cellcolor{blue!10}Tk-Instruct \citep{super} & 41.3 & 41.8 & 42.2 & 38.9 & - & - & - & 41.4 & 41.8 & 42.1 & 41.5 & 41.8 & 40.6 \\
        \cellcolor{blue!10}\textbf{TAGI (Ours)} & 42.1 & 42.5 & 40.3 & 39.7 & 41.8 & 42.5 & 42.3 & 41.8 & 42.5 & 42.4 & 42.3 & 42.5 & 41.9 \\ \hline
        \rowcolor{pink!40}\multicolumn{14}{l}{\textit{Def}} \\ \hline
        \cellcolor{pink!40}Tk-Instruct \citep{super} & 35.0 & 34.2 & 32.6 & 31.7 & - & - & - & 34.4 & 34.2 & 34.5 & 35.0 & 34.2 & 34.3\\
        \cellcolor{pink!40}\textbf{TAGI (Ours)} & 34.3 & 35.3 & 33.5 & 31.8 & 34.8 & 35.3 & 35.4 & 34.2 & 35.3 & 35.4 & 34.8 & 35.3 & 34.9 \\ \hline
        \rowcolor{green!10}\multicolumn{14}{l}{\textit{\textbf{P3}}} \\ 
        \rowcolor{yellow!30}\multicolumn{14}{l}{\textit{MTest11 Avg.}} \\ \hline
        \cellcolor{yellow!30}Metatrain & 43.3 & 44.1 & 43.6 & 40.9 & - & - & - & 44.0 & 44.1 & 44.3 & 44.2 & 44.1 & 43.6 \\
        \cellcolor{yellow!30}\textbf{TAGI (Ours)} & 44.0 & 45.6 & 44.0 & 41.6 & 44.8 & 45.6 & 45.5 & 44.3 & 45.6 & 45.2 & 45.1 & 45.6 & 44.8 \\
        \bottomrule
      \end{tabular}
      }
      % }
\end{table}

\subsection{How $\lambda_1$ and $\lambda_2$ are tuned?}

\begin{figure}[t]
\centerline{\includegraphics[width=1.0\textwidth]{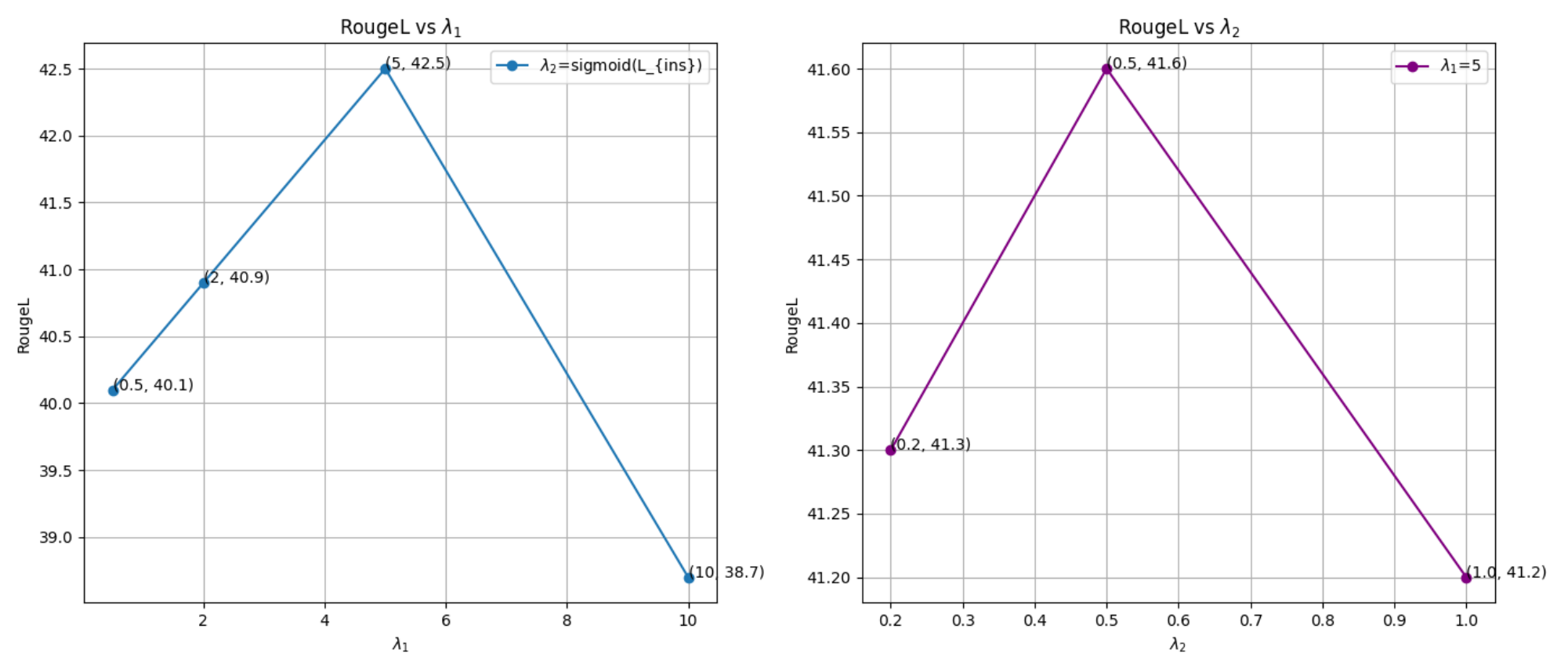}}
% \vspace{-0.45cm}
\caption{Ablation study on $\lambda$ hyperparameters. The backbone model is T5-Base.
}
\label{a}
% \vspace{-0.65cm}
\end{figure}

In the experiment, we set $\lambda_1$ and $\lambda_2$ to two different values: $\lambda_1 = 5$ and $\lambda_2 = \text{sigmoid}(\mathcal{L}_{\text{ins}})$. The effects of these different $\lambda$ values on the results are illustrated in Figure \ref{a} and Table \ref{table:a}. We maintained all other conditions constant and only varied $\lambda$ to perform an ablation experiment at Def+2Pos. scenario.

\begin{table}[h]
\centering
\caption{Ablation study on $\lambda$ hyperparameters. The backbone model is T5-Base.}
\begin{tabular}{ccc}
\hline
 $\lambda_1$ & $\lambda_2$ & RougeL \\
\hline
 0.5 & $sigmoid(\mathcal{L}_{\text{ins}})$ & 40.1 \\

  2 & $sigmoid(\mathcal{L}_{\text{ins}})$ & 40.9 \\

  5 & $sigmoid(\mathcal{L}_{\text{ins}})$ & \textbf{42.5} \\

  10 & $sigmoid(\mathcal{L}_{\text{ins}})$ & 38.7 \\
\hline
  5 & 0.2 & 41.3 \\

 5 & 0.5 & 41.6 \\

 5 & 1.0 & 41.2 \\
\hline
\end{tabular}

\label{table:a}
\end{table}

\subsection{Inference Cost}

To analyze the computational efficiency of the TAGI model compared to the standard instruction training model (full fine-tuning), let's consider a scenario where we have to process \( n \) samples, each of length \( i \), along with a task instruction of length \( t \). We assume the output sequence length is negligible and thus ignore it in our computations.

In a typical full fine-tuning setup, such as Tk-Instruct, each input is concatenated with the task instruction, requiring the model to process the combined input sequence. If we denote the number of FLOPs required to process a single token with an encoder-decoder model as \( N \), where \( N \) is the total number of model parameters, then the total computation cost for all samples can be estimated as:
$ \text{FLOPs}_{\text{standard}} = N \cdot n(t + i) $
Here, each of the \( n \) samples includes both the instruction and the sample input, leading to \( n(t + i) \) tokens being processed.

Our TAGI model, on the other hand, processes the task instruction only once, regardless of the number of samples. This unique feature significantly reduces the computation required, especially as the number of samples or the length of the instruction increases. The total computation cost in this model is given by:
$ \text{FLOPs}_{\text{TAGI}} = N \cdot (t + ni) $
In this case, the instruction length \( t \) is processed only once, and each sample is processed separately, resulting in a total of \( (t + ni) \) tokens being processed.

\section{Extended Results}
\subsection{Characteristics Comparison of the Proposed TAGI and Other Baselines}
\label{sec:c_full}
Here, we report a full comparison of methods and the proposed TAGI in \autoref{comparison}, also visualized in \autoref{c_half}. In this report, we compare various methods across eight dimensions. \textbf{Finetuning on target tasks} yields good performance; however, it necessitates retuning when applied to unseen tasks and fails to address these effectively. \textbf{Strong baseline meta-training methods} excel at handling unseen tasks by enabling models to solve problems based on task-specific instructions. Nevertheless, these methods are limited to instance-level operations and entail repetitive processing of concatenated instructions and comprehensive finetuning, resulting in significant parameter updates and high inference costs.

\textbf{Hypter} \citep{ye2021learning} initially introduced the approach of considering tasks at the task level, treating identical tasks as a unified entity, and employing a hypernetwork to generate adapters that represent specific task models from instructions. Building on this, \textbf{Hypertuning} \citep{hypertuning} uses demonstrations to generate adapters and pretrains the hypernetwork to boost its expressive capabilities. Both strategies avoid the direct input of instructions and rely on hypernetwork, which reduces parameter updates and lowers computational demands during inference. However, they suffer from notable performance degradation due to the lack of instructional information in the input.

\textbf{HINT} \citep{hint} addresses this issue by appending instructions post-encoder, thus eliminating redundant computations. Although these methods facilitate learning at the task level, they do not engage in instruction-based learning, i.e., they do not explicitly supervise the hypernetwork's generation process to aid in understanding instructions and generating parameters.

\textbf{The proposed TAGI} rectifies these deficiencies by integrating cross-attention for enhanced information fusion and supervised learning of adapter weights within HINT. This innovation aids in generalizing to unseen tasks without increasing the computational burden.
\begin{table}[htbp]
  \caption{Compare the characteristics of all comparison methods and the proposed TAGI.}
  \label{comparison}
  \centering
  \renewcommand\arraystretch{1.05}
  \resizebox{\linewidth}{!}{
      \begin{tabular}{lcccccccc}
        \toprule
                & Meta-  & Pre-  & Instr.  & Instr. & Low Up. & Low Infer. & Instr.    & Unseen  \\
        Method  & Train  & Train & Concat. & Fus.   & Params  & Cost       & Learning  & Task \\
        \midrule
        Simple FT & \XSolidBrush  & \XSolidBrush & \CheckmarkBold &  \CheckmarkBold & \XSolidBrush & \XSolidBrush  & \XSolidBrush & \XSolidBrush \\
        T0 \citep{multitask} / Tk-Instruct \citep{super}   & \CheckmarkBold & \XSolidBrush & \CheckmarkBold & \CheckmarkBold & \XSolidBrush & \XSolidBrush & \XSolidBrush & \CheckmarkBold\CheckmarkBold\CheckmarkBold   \\
        Hypter \citep{ye2021learning}   & \CheckmarkBold & \XSolidBrush & \XSolidBrush & \XSolidBrush &  \CheckmarkBold &  \CheckmarkBold & \XSolidBrush  &  \CheckmarkBold  \\
        HyperTuning \citep{hypertuning}  & \CheckmarkBold & \CheckmarkBold & \XSolidBrush & \XSolidBrush &  \CheckmarkBold &  \CheckmarkBold & \XSolidBrush  & \CheckmarkBold  \\
        HINT \citep{hint} & \CheckmarkBold & \CheckmarkBold & \CheckmarkBold & \XSolidBrush &  \CheckmarkBold &  \CheckmarkBold & \XSolidBrush  &  \CheckmarkBold\CheckmarkBold  \\
        \textbf{TAGI (Ours)}   & \CheckmarkBold   & \CheckmarkBold & \CheckmarkBold & \CheckmarkBold & \CheckmarkBold & \CheckmarkBold & \CheckmarkBold   & \CheckmarkBold\CheckmarkBold\CheckmarkBold \\
        \bottomrule
      \end{tabular}
      }
\end{table}

\subsection{P3 Full Results}
\label{sec:exres}

\autoref{p3_main} reports the per-task performance and average accuracy on P3 reported in \autoref{p3_t}.

\begin{table}[htbp]
  \caption{Main Full P3 Results. "-" means not reported. $\dagger$ and $\ddagger$ mean the results are from FiD-ICL \citep{ye-etal-2023-fid} and Hypertuning \citep{hypertuning} respectively. $\Diamond$ Computed as the average of R1/R2/R3 (except for HyperT5 rows where the numbers are quoted). More ICL-based results and details can be seen FiD-ICL \citep{ye-etal-2023-fid}.}
  \label{p3_main}
  \centering
  \renewcommand\arraystretch{1.1}
  \resizebox{\linewidth}{!}{
      \begin{tabular}{lcccccccccccc|C{3.5em}C{4em}}
        \toprule
        % & \multicolumn{3}{c}{\textbf{T5-LM}} & \multicolumn{3}{c}{\textbf{T0}}  \\
        % \cmidrule(r){2-4}\cmidrule(r){5-7}
        Method & ANLI $^\Diamond$ & (R1) & (R2) & (R3) & HSwag & CB & COPA & RTE & WiC & WSC & WGD & SCloze & MTest11 Avg. & HyperT5 Avg. \\
        \hline
        Random & 33.4 & 33.4 & 33.4 & 33.4 & 25.0 & 50.0 & 50.0 & 52.7 & 50.0 & 63.5 & 50.0 & 50.0 & 44.7 & 46.8 \\
        \hline
        \rowcolor{gray!20} \multicolumn{15}{l}{\textit{\# Base(250M)}}\\ \hline
        T5-LM $^\dagger$ & 33.4 & 33.3 & 33.5 & 33.5 & 24.7 & 44.3 & 54.3 & 47.9 & 49.7 & 57.9 & 49.8 & 54.1 & 43.9  & 45.2 \\
        \cellcolor{blue!10}T5-LM Full FT $^\dagger$ & 33.8 & 34.5 & 33.4 & 33.5 & 24.8 & 66.5 & 45.7 & 51.1 & 53.7 & 46.3 & 49.8 & 50.9 & 44.6 & 46.5 \\
        \cellcolor{blue!10}T5-LM Metatrain & 31.0 & 30.3 & 29.5 & 33.1 & 25.0 & 40.5 & 52.6 & 51.2 & 50.2 & 58.4 & 47.4 & 66.6 & 44.1 & 44.6 \\
        \cellcolor{orange!10}T5-LM-FiD $^\dagger$ & 33.0 & 32.4 & 33.1 & 33.4 & 26.7 & 42.5 & 58.8 & 54.6 & 51.1 & 57.9 & 50.3 & 76.3 & 47.0  & 46.9 \\
        \cellcolor{green!10}\textbf{T5-LM-TAGI} & 32.1 & 31.5 & 31.7 & 33.1 & 25.0 & 44.5 & 54.7 & 53.7 & 52.3 & 60.5 & 50.8 & 64.0 & 45.6 & 46.7 \\
        \hline
        T0 $^\dagger$ & 32.3 & 31.5 & 32.4 & 33.1 & 26.5 & 45.8 & 65.9 & 69.3 & 51.6 & 56.7 & 51.2 & 76.1 & 49.1 & 49.9 \\
        \cellcolor{blue!10}T0 Full FT $^\dagger$ & 33.5 & 32.6 & 33.9 & 33.9 & 29.1 & 73.2 & 66.3 & 68.0 & 53.1 & 50.9 & 51.0 & 79.0 & 51.9  & 53.1 \\
        \cellcolor{blue!10}T0 Metatrain & 32.1 & 31.5 & 31.5 & 33.2 & 29.5 & 50.4 & 64.2 & 68.2 & 47.7 & 61.6 & 52.8 & 80.8 & 50.1 & 50.8 \\
        \cellcolor{orange!10}T0-FiD $^\dagger$ & 32.7 & 31.7 & 32.9 & 33.6 & 26.2 & 54.9 & 68.2 & 68.1 & 51.9 & 60.3 & 51.3 & 82.3 & 51.0 & 51.7 \\
        \cellcolor{green!10}\textbf{T0-TAGI} & 32.7 & 31.1 & 31.9 & 35.0 & 29.8 & 49.3 & 67.1 & 70.0 & 49.0 & 61.2 & 54.4 & 79.6 & 50.8 & 51.7 \\
         \hline
        \rowcolor{gray!20} \multicolumn{15}{l}{\textit{\# Large(800M)}}\\ \hline
        T5-LM $^\dagger$ & 32.7 & 32.1 & 33.4 & 32.7 & 25.3 & 33.8 & 50.5 & 49.0 & 51.0 & 50.4 & 50.5 & 47.8 & 41.5 & 42.9 \\
        \cellcolor{blue!10}T5-LM Full FT $^\dagger$ & 34.1 & 35.1 & 33.6 & 33.6 & 26.1 & 65.4 & 47.1 & 51.7 & 53.5 & 47.5 & 49.9 & 56.5 & 45.5  & 46.9 \\
        \cellcolor{blue!10}T5-LM Metatrain & 31.3 & 30.0 & 30.5 & 33.4 & 27.0 & 60.4 & 77.6 & 71.9 & 47.0 & 56.4 & 54.8 & 87.2 & 52.4 & 53.3 \\
        \cellcolor{orange!10}T5-LM-FiD $^\dagger$ & 34.4 & 33.9 & 33.4 & 35.8 & 28.3 & 60.2 & 81.1 & 72.6 & 50.7 & 63.7 & 55.6 & 91.6 & 55.2  & 55.8 \\
        \cellcolor{green!10}\textbf{T5-LM-TAGI} & 33.7 & 33.5 & 32.5 & 35.1 & 27.8 & 62.9 & 79.0 & 76.1 & 52.9 & 57.9 & 58.2 & 86.2 & 54.7 & 56.0 \\
        \hline
        T0 $^\dagger$ & 34.1 & 32.2 & 34.2 & 36.0 & 26.1 & 56.8 & 76.6 & 65.3 & 50.8 & 56.4 & 53.9 & 88.4 & 52.4 & 52.5 \\
        \cellcolor{blue!10}T0 Full FT $^\dagger$ & 35.3 & 34.5 & 35.4 & 36.2 & 33.1 & 80.1 & 80.8 & 69.2 & 54.1 & 53.2 & 56.3 & 90.0 & 56.6  & 57.8 \\
        \cellcolor{blue!10}T0 Metatrain & 32.9 & 31.5 & 31.8 & 35.5 & 24.5 & 59.4 & 77.0 & 65.1 & 48.8 & 56.7 & 57.6 & 88.0 & 52.4 & 52.8 \\
        \cellcolor{orange!10}T0-FiD $^\dagger$ & 33.4 & 31.8 & 32.8 & 35.7 & 26.1 & 60.7 & 77.6 & 67.1 & 52.1 & 59.1 & 54.7 & 89.5 & 53.4  & 53.9 \\
        \cellcolor{green!10}\textbf{T0-TAGI} & 32.7 & 31.5 & 32.9 & 36.6 & 27.3 & 61.3 & 79.6 & 68.7 & 48.2 & 59.9 & 56.4 & 89.4 & 53.8 & 54.6 \\
        \hline
        \cellcolor{pink!30}HyperT5-Prefix $^\ddagger$ & 33.4 & - & - & - & 32.3 & 60.1 & 73.9 & 71.5 & 51.1 & 63.0 & 51.1 & - & -  & 54.6 \\
        \cellcolor{pink!30}HyperT5-LoRA $^\ddagger$ & 33.6 & - & - & - & 33.0 & 49.5 & 74.2 & 67.4 & 52.0 & 64.0 & 52.9 & - & -  & 53.3 \\
         \hline
        \rowcolor{gray!20} \multicolumn{15}{l}{\textit{\# XL(3B)}}\\ \hline
        T5-LM $^\dagger$ & 32.7 & 32.2 & 33.4 & 32.7 & 24.6 & 32.7 & 53.1 & 48.8 & 50.8 & 57.6 & 50.9 & 51.4 & 42.6 & 43.9 \\
        \cellcolor{blue!10}T5-LM Full FT $^\dagger$ & 34.6 & 35.5 & 34.3 & 33.9 & 27.1 & 67.8 & 54.8 & 50.7 & 53.7 & 47.7 & 50.7 & 63.3 & 47.2  & 48.4 \\
        \cellcolor{blue!10}T5-LM Metatrain & 32.7 & 31.5 & 32.3 & 34.3 & 33.3 & 59.5 & 74.8 & 69.5 & 52.6 & 53.8 & 54.2 & 88.4 & 53.1 & 53.8 \\
        \cellcolor{orange!10}T5-LM-FiD $^\dagger$ & 39.3 & 39.8 & 37.6 & 40.4 & 31.4 & 67.0 & 92.3 & 78.8 & 50.4 & 64.5 & 61.2 & 96.5 & 60.0 & 60.6 \\
        \cellcolor{green!10}\textbf{T5-LM-TAGI} & 37.7 & 37.8 & 36.1 & 39.3 & 32.0 & 68.2 & 89.4 & 76.6 & 53.6 & 61.2 & 59.6 & 94.2 & 58.9 & 59.8 \\
        \hline
        T0 $^\dagger$ & 38.0 & 38.4 & 35.7 & 40.0 & 26.5 & 67.7 & 82.2 & 80.1 & 53.5 & 57.3 & 57.8 & 94.0 & 57.6  & 57.9 \\
        \cellcolor{blue!10}T0 Full FT $^\dagger$ & 38.5 & 37.5 & 38.8 & 39.2 & 38.7 & 81.9 & 88.0 & 80.1 & 55.9 & 59.5 & 61.4 & 95.0 & 61.4 & 63.0 \\
        \cellcolor{blue!10}T0 Metatrain & 37.0 & 37.3 & 33.2 & 40.4 & 24.8 & 66.9 & 81.9 & 78.9 & 52.7 & 60.2 & 55.6 & 92.8 & 56.8 & 57.3 \\
        \cellcolor{orange!10}T0-FiD $^\dagger$ & 38.6 & 39.0 & 36.5 & 40.5 & 28.5 & 62.9 & 87.4 & 74.6 & 52.1 & 62.7 & 61.0 & 95.5 & 58.2 & 58.5 \\
        \cellcolor{green!10}\textbf{T0-TAGI} & 38.7 & 39.5 & 35.6 & 41.0 & 26.5 & 68.7 & 87.8 & 78.2 & 52.2 & 61.8 & 59.8 & 95.6 & 58.8 & 59.2 \\
        \hline
        \cellcolor{pink!30}HyperT5-Prefix $^\ddagger$ & 38.7 & - & - & - & 33.6 & 69.6 & 88.4 & 79.5 & 53.1 & 57.6 & 56.6 & - & -  & 59.6 \\
        \cellcolor{pink!30}HyperT5-LoRA $^\ddagger$ & 35.3 & - & - & - & 30.8 & 66.4 & 83.3 & 68.5 & 50.3 & 60.0 & 56.1 & - & -  & 56.4 \\
        \bottomrule
      \end{tabular}
      }
\end{table}

\section{Limitations}
\label{limit}
\noindent \textbf{Large Language Models.} Due to computational constraints, most of our experiments were conducted using models with $\leq3B$ parameters. Given the complexity of our research, we restricted our focus to encoder-decoder models, which have demonstrated superior performance in cross-task generalization \citep{language}, which we explore further in \ref{sec:encvsdec}. Consequently, it remains uncertain whether instruction learning can be effectively scaled to larger models ($\geq 7B$ parameters) or commonly used decoder-only models. However, since our method preserves the original model parameters without compromising performance, we anticipate its applicability to broader research in the future.

\noindent \textbf{Training Costs.} Although TAGI is computationally efficient during inference, its training cost is significantly higher. This is due to the additional requirements beyond the foundation laid by previous work, including the introduction of knowledge distillation, running a hypernetwork to generate adapters for each batch, and pre-training some downstream task-specific models. Consequently, while TAGI may be highly efficient for inference and suitable for users with limited resources, training a unique TAGI model presents considerable challenges.

\noindent \textbf{Datasets.} In the SNI study, our investigation was limited to tasks in English, leaving the generalization capabilities in a multilingual context unexplored. However, given the proven effectiveness of hypernetwork methods in achieving multilingual generalization \citep{multilingual, hyperx}, we are optimistic about the potential directions for our future research in this domain. Furthermore, in P3, we adopted the methodologies of T0 \citep{multitask} and FiD-ICL \citep{ye-etal-2023-fid}, concentrating primarily on natural language processing (NLP) tasks amenable to ranking classification. This focus included tasks related to classification and multiple-choice questions but excluded other types of generative tasks. Looking ahead, we aim to develop new research resources and broaden our experimental scope and evaluations to encompass a more diverse array of categories.

\begin{minipage}{\textwidth}
\begin{minipage}[t]{0.48\textwidth}
\makeatletter\def\@captype{table}
\centering
\caption{Meta-Train dataset of Super-Natural Instructions.}
\resizebox{\linewidth}{!}{
\begin{tabular}{lc}
    \toprule
    Task     & \# Num of Task  \\
    \midrule
    \rowcolor{green!10} \multicolumn{2}{l}{First Split (6 Tasks)}\\
    Question Answering & 157\\
    Program Execution & 90 \\
    Question Generation & 51 \\
    Sentiment Analysis & 42 \\
    Misc. & 36 \\
    Toxic Language Detection & 32 \\ \hline
    \rowcolor{green!10} \multicolumn{2}{l}{Second Split (15 Tasks)}\\
    Text Categorization & 28 \\
    Commonsense Classification & 23 \\
    Text Matching & 17 \\
    Named Entity Recognition & 17 \\
    Information Extraction & 17 \\
    Wrong Candidate Generation & 15 \\
    Text Completion & 14 \\
    Question Understanding & 13 \\
    Text to Code & 12 \\ \hline
    \rowcolor{green!10} \multicolumn{2}{l}{Third Split (30 Tasks)}\\
    Summarization & 12 \\
    Dialogue Generation & 11 \\
    Word Semantics & 10 \\
    Story Composition & 9 \\
    Speaker Identification & 9 \\
    Pos Tagging & 9 \\
    Linguistic Probing & 9 \\
    Fill in The Blank & 8 \\
    Text Quality Evaluation & 7 \\
    Stereotype Detection & 7 \\
    Sentence Composition & 7 \\
    Negotiation Strategy Detection & 7 \\
    Gender Classification & 7 \\
    Coherence Classification & 6 \\
    Word Relation Classification & 5 \\ \hline
    \rowcolor{green!10} \multicolumn{2}{l}{Fourth Split (60 Tasks)}\\
    Explanation & 5 \\
    Text Simplification & 4 \\
    Sentence Perturbation & 4 \\
    Paraphrasing & 4 \\
    Mathematics & 4 \\
    Intent Identification & 4 \\
    Dialogue State Tracking & 4 \\
    Code to Text & 4 \\
    Sentence Ordering & 3 \\
    Fact Verification & 3 \\
    Answer Verification & 3 \\
    Translation & 2 \\
    Style Transfer & 2 \\
    Stance Detection & 2 \\
    Speaker Relation Classification & 2 \\
    Question Decomposition & 2 \\
    Number Conversion & 2 \\
    Irony Detection & 2 \\
    Grammar Error Detection & 2 \\
    Spelling Error Detection & 1 \\
    Spam Classification & 1 \\
    Sentence Expansion & 1 \\
    Sentence Compression & 1 \\
    Punctuation Error Detection & 1 \\
    Preposition Prediction & 1 \\
    Poem Generation & 1 \\
    Entity Relation Classification & 1 \\
    Entity Generation & 1 \\
    Discourse Relation Classification & 1 \\
    Discourse Connective Identification & 1 \\ 
    \bottomrule
\end{tabular}
}

\label{sni_list}
\end{minipage}
\begin{minipage}[t]{0.48\textwidth}
\makeatletter\def\@captype{table}
\centering
\caption{P3 dataset tasks. $\dagger$ means evaluation without story\_cloze.}
\resizebox{\linewidth}{!}{
\begin{tabular}{lc}
    \toprule
    Task   & \# Num of Prompts  \\
    \midrule
    \rowcolor{gray!20} \multicolumn{2}{l}{Meta-Train}\\ \hline
    \rowcolor{green!10} \multicolumn{2}{l}{First Split (5 Tasks)}\\
    cosmos\_qa & 13 \\ 
    kilt\_tasks\_hotpotqa & 5 \\ 
    amazon\_polarity & 9 \\ 
    cnn\_dailymail\_3.0.0 & 9 \\ 
    common\_gen & 9 \\ \hline
    \rowcolor{green!10} \multicolumn{2}{l}{Second Split (10 Tasks)}\\
    glue\_mrpc & 7\\ 
    adversarial\_qa\_dbert & 5 \\
    ag\_news & 7\\ 
    dream & 5\\   
    gigaword & 9\\ \hline
    \rowcolor{green!10} \multicolumn{2}{l}{Third Split (20 Tasks)}\\
    paws & 12\\ 
    wiki\_qa & 11 \\ 
    ropes & 12 \\ 
    quoref & 11\\ 
    dbpedia\_14 & 4\\
    multi\_news & 6 \\
    imdb & 10 \\
    quail & 13\\ 
    quartz & 8\\ 
    wiki\_bio & 5\\ \hline
    \rowcolor{green!10} \multicolumn{2}{l}{Fourth Split (36 Tasks)}\\
    adversarial\_qa\_dbidaf & 5 \\ 
    adversarial\_qa\_droberta  & 5\\ 
    duorc\_SelfRC & 9\\
    duorc\_ParaphraseRC & 9\\
    cos\_e\_v1.11 & 11\\ 
    qasc & 8 \\ 
    sciq & 5 \\ 
    glue\_qqp & 6\\ 
    social\_i\_qa & 6\\ 
    wiki\_hop\_original & 9 \\ 
    wiqa & 8\\ 
    app\_reviews & 4 \\ 
    rotten\_tomatoes & 10 \\ 
    yelp\_review\_full & 7 \\ 
    samsum & 7 \\ 
    xsum & 10\\ 
     \hline
    \rowcolor{gray!20} \multicolumn{2}{l}{Meta-Test}\\ \hline
    super\_glue\_wsc.fixed  \\
    winogrande\_winogrande\_xl \\
    super\_glue\_cb \\
    super\_glue\_rte  \\
    anli(r1/r2/r3)  \\
    super\_glue\_copa \\
    hellaswag \\
    super\_glue\_wic  \\
    story\_cloze $^\dagger$ \\
    \bottomrule
\end{tabular}

}
\label{p3_list}
\end{minipage}
\end{minipage}

%%%%%%%%%%%%%%%%%%%%%%%%%%%%%%%%%%%%%%%%%%%%%%%%%%%%%%%%%%%%
%%%%%%%%%%%%%%%%%%%%%%%%%%%%%%%%%%%%%%%%%%%%%%%%%%%%%%%%%%%%

\newpage
\section*{NeurIPS Paper Checklist}

\begin{enumerate}

\item {\bf Claims}
    \item[] Question: Do the main claims made in the abstract and introduction accurately reflect the paper's contributions and scope?
    \item[] Answer: \answerYes{} % Replace by \answerYes{}, \answerNo{}, or \answerNA{}.
    \item[] Justification: The abstract and introduction accurately reflect the paper's contributions and scope.

\item {\bf Limitations}
    \item[] Question: Does the paper discuss the limitations of the work performed by the authors?
    \item[] Answer: \answerYes{} % Replace by \answerYes{}, \answerNo{}, or \answerNA{}.
    \item[] Justification: We can find the limitations in \ref{limit}.

\item {\bf Theory Assumptions and Proofs}
    \item[] Question: For each theoretical result, does the paper provide the full set of assumptions and a complete (and correct) proof?
    \item[] Answer: \answerNA{} % Replace by \answerYes{}, \answerNo{}, or \answerNA{}.
    \item[] Justification:  Our paper does not include theoretical results.

    \item {\bf Experimental Result Reproducibility}
    \item[] Question: Does the paper fully disclose all the information needed to reproduce the main experimental results of the paper to the extent that it affects the main claims and/or conclusions of the paper (regardless of whether the code and data are provided or not)?
    \item[] Answer: \answerYes{} % Replace by \answerYes{}, \answerNo{}, or \answerNA{}.
    \item[] Justification: We can reproduce the main experimental results following our settings in \ref{experiment} and \ref{main_ex}.

\item {\bf Open access to data and code}
    \item[] Question: Does the paper provide open access to the data and code, with sufficient instructions to faithfully reproduce the main experimental results, as described in supplemental material?
    \item[] Answer: \answerYes{} % Replace by \answerYes{}, \answerNo{}, or \answerNA{}.
    \item[] Justification: We'll open source the code to an anonymous site \url{https://anonymous.4open.science/r/TAGI} and put it on github after review.

\item {\bf Experimental Setting/Details}
    \item[] Question: Does the paper specify all the training and test details (e.g., data splits, hyperparameters, how they were chosen, type of optimizer, etc.) necessary to understand the results?
    \item[] Answer: \answerYes{} % Replace by \answerYes{}, \answerNo{}, or \answerNA{}.
    \item[] Justification: We can find the experimental settings (hyperparameters and datasets) in \ref{ex_im} and \ref{experiment}.

\item {\bf Experiment Statistical Significance}
    \item[] Question: Does the paper report error bars suitably and correctly defined or other appropriate information about the statistical significance of the experiments?
    \item[] Answer: \answerYes{} % Replace by \answerYes{}, \answerNo{}, or \answerNA{}.
    \item[] Justification: We examined the effect of different hyperparameters on results in \ref{sec:analysis}.

\item {\bf Experiments Compute Resources}
    \item[] Question: For each experiment, does the paper provide sufficient information on the computer resources (type of compute workers, memory, time of execution) needed to reproduce the experiments?
    \item[] Answer: \answerYes{} % Replace by \answerYes{}, \answerNo{}, or \answerNA{}.
    \item[] Justification: We can find it in \ref{ex_im} and \ref{experiment}.
    
\item {\bf Code Of Ethics}
    \item[] Question: Does the research conducted in the paper conform, in every respect, with the NeurIPS Code of Ethics \url{https://neurips.cc/public/EthicsGuidelines}?
    \item[] Answer: \answerYes{} % Replace by \answerYes{}, \answerNo{}, or \answerNA{}.
    \item[] Justification: All of our studies follow the NeurIPS Code of Ethics.

\item {\bf Broader Impacts}
    \item[] Question: Does the paper discuss both potential positive societal impacts and negative societal impacts of the work performed?
    \item[] Answer: \answerNA{} % Replace by \answerYes{}, \answerNo{}, or \answerNA{}.
    \item[] Justification: There is no societal impact of the work performed.
    
\item {\bf Safeguards}
    \item[] Question: Does the paper describe safeguards that have been put in place for responsible release of data or models that have a high risk for misuse (e.g., pretrained language models, image generators, or scraped datasets)?
    \item[] Answer: \answerNA{} % Replace by \answerYes{}, \answerNo{}, or \answerNA{}.
    \item[] Justification: The paper poses no such risks.

\item {\bf Licenses for existing assets}
    \item[] Question: Are the creators or original owners of assets (e.g., code, data, models), used in the paper, properly credited and are the license and terms of use explicitly mentioned and properly respected?
    \item[] Answer: \answerYes{} % Replace by \answerYes{}, \answerNo{}, or \answerNA{}.
    \item[] Justification: We follow their open-source protocols in all our uses.

\item {\bf New Assets}
    \item[] Question: Are new assets introduced in the paper well documented and is the documentation provided alongside the assets?
    \item[] Answer: \answerNA{} % Replace by \answerYes{}, \answerNo{}, or \answerNA{}.
    \item[] Justification: This paper does not release new assets.

\item {\bf Crowdsourcing and Research with Human Subjects}
    \item[] Question: For crowdsourcing experiments and research with human subjects, does the paper include the full text of instructions given to participants and screenshots, if applicable, as well as details about compensation (if any)? 
    \item[] Answer: \answerNA{} % Replace by \answerYes{}, \answerNo{}, or \answerNA{}.
    \item[] Justification: This paper does not involve crowdsourcing nor research with human subjects.

\item {\bf Institutional Review Board (IRB) Approvals or Equivalent for Research with Human Subjects}
    \item[] Question: Does the paper describe potential risks incurred by study participants, whether such risks were disclosed to the subjects, and whether Institutional Review Board (IRB) approvals (or an equivalent approval/review based on the requirements of your country or institution) were obtained?
    \item[] Answer: \answerNA{} % Replace by \answerYes{}, \answerNo{}, or \answerNA{}.
    \item[] Justification: This paper does not involve crowdsourcing nor research with human subjects.

\end{enumerate}

\end{document}